%% file: acl_latex.tex
\documentclass[11pt]{article}

\usepackage[final]{acl}

\usepackage{times}
\usepackage{latexsym}

\usepackage[T1]{fontenc}
\usepackage[utf8]{inputenc}

\usepackage{microtype}

\usepackage{inconsolata}

\usepackage{graphicx}
\usepackage{dblfloatfix}
\usepackage{placeins}
\usepackage{pgfplots}
\pgfplotsset{compat=1.18}

\usepackage{amsmath}
\usepackage{amssymb}
\usepackage{array}
\usepackage{booktabs}
\usepackage{tabularx}
\usepackage{multirow}
\usepackage{tcolorbox}
\usepackage{listings}
\usepackage{colortbl}
\tcbuselibrary{breakable}

\definecolor{ourrow}{RGB}{230,240,250}
\definecolor{tblCtrlBg}{RGB}{245,247,250}
\definecolor{tblAtrBg}{RGB}{232,242,251}
\definecolor{tblAlwBg}{RGB}{255,240,220}
\definecolor{tblGapBg}{RGB}{239,242,246}
\definecolor{tblBestBg}{RGB}{255,226,166}
\definecolor{tblAtrText}{RGB}{44,109,170}
\definecolor{tblAlwText}{RGB}{190,95,0}
\definecolor{tblMutedText}{RGB}{139,148,158}

\newcolumntype{Y}{>{\centering\arraybackslash}X}
\newcommand{\xmark}{\ensuremath{\times}}
\newcommand{\bestcell}[1]{\cellcolor{tblBestBg}\textbf{#1}}
\newcommand{\sparsecell}[1]{\textcolor{tblMutedText}{#1}}
\newcommand{\defhdr}{\textcolor{tblMutedText}{\textbf{Def.}}}
\newcommand{\atrhdr}{\textcolor{tblAtrText}{\textbf{ATR}}}
\newcommand{\alwhdr}{\textcolor{tblAlwText}{\textbf{Alw.}}}
\newcommand{\orchdr}{\textbf{Orc.}}

\title{Ask Now, Use Later: Benchmarking the Proactivity Gap in Long-Lived LLM Agents}

\author{
\textbf{Bin Wu\textsuperscript{1,*}},
\textbf{Guanyun Zou\textsuperscript{2,*}},
\textbf{Bingbing Wang\textsuperscript{3,*}},
\textbf{Huan Zhao\textsuperscript{4}},
\textbf{Chuan Shi\textsuperscript{1,$\dagger$}}
\\
\normalfont
\textsuperscript{1}Beijing University of Posts and Telecommunications
\\
\textsuperscript{2}Nanjing University of Aeronautics and Astronautics
\\
\textsuperscript{3}Harbin Institute of Technology (Shenzhen)
\\
\textsuperscript{4}Noumena AI
\\
\texttt{wb789@bupt.edu.cn},
\texttt{zouguanyun@nuaa.edu.cn},
\texttt{bingbing.wang@stu.hit.edu.cn}
\\
\texttt{zhaohuan@noumena.com.cn},
\texttt{shichuan@bupt.edu.cn}
}

\newcommand{\authornotetext}{%
  \begingroup
  \renewcommand{\thefootnote}{\fnsymbol{footnote}}%
  \footnotetext[1]{Work done during internships at Noumena AI.}%
  \footnotetext[2]{Corresponding author.}%
  \endgroup
}

\begin{document}
\maketitle
\authornotetext
\begin{abstract}
A long-lived LLM agent, such as OpenClaw, earns its value by acting on a user's preferences and constraints across sessions, not just the current request. Yet today's agents keep what a user volunteers but rarely ask for what stays unspoken, leaving a \textbf{proactivity gap in long-lived LLM agents}: an agent cannot act on a preference it never obtained. As users delegate more of their affairs to agents, the impact of this gap grows. We isolate one concrete, controllable slice of this gap as \textbf{Ask-to-Remember (ATR)}: the agent decides whether to ask now for a reusable user preference that the current task does not need but a later session with the same user will. ATR is hard even to evaluate: the right question is underdetermined and its payoff deferred to tasks that may never arise. \textbf{ATRBench}, to the best of our knowledge the first ATR benchmark, makes it measurable by fixing each user's preferences as hidden ground truth, so success demands asking, not recall. Across eight frontier LLM agents, defaults fall at least 62 points below an oracle handed the relevant preference, and prompting closes little of it. Diagnostics identify acquisition as the bottleneck. ATRBench surfaces this proactivity gap in current agents and offers a diagnostic testbed for closing it.\footnote{Code and released benchmark data: \url{https://github.com/BUPT-GAMMA/ATRBench}}
\end{abstract}

% =========================================================
% PAPER CONTENT
% =========================================================

\input{sections/intro}

\input{sections/problem}

\input{sections/benchmark}

\input{sections/experiments}

\input{sections/related}

\input{sections/conclusion}

\input{sections/limitations}

\input{sections/ethics}

\input{sections/acknowledgments}

\bibliography{custom}

\clearpage
\appendix
\raggedbottom
\input{sections/appendix}

\end{document}

%% file: sections/intro.tex
\section{Introduction}
\label{sec:intro}

LLM agents are increasingly developed as long-lived personal assistants that serve the same user across many sessions, from research systems~\citep{park2023generative,packer2023memgpt,zhong2024memorybank,brady2026springdrift} to open-source runtimes such as OpenClaw.\footnote{\url{https://openclaw.ai/}} The value of such an assistant lies not in handling any single request in isolation, but in accumulating user information over time and acting on it in later tasks. Such information includes what the user prefers, what standing constraints they hold, and how they want tasks carried out. As these agents take on recurring matters on the user's behalf, such as scheduling, travel, communication, and shopping, personalization turns from a convenience into a precondition for reliable action. An assistant that does not know the user's standing preferences may carry out a request to the letter and still act in a way the user would not accept.

\begin{figure}[t]
  \centering
  \includegraphics[width=\columnwidth]{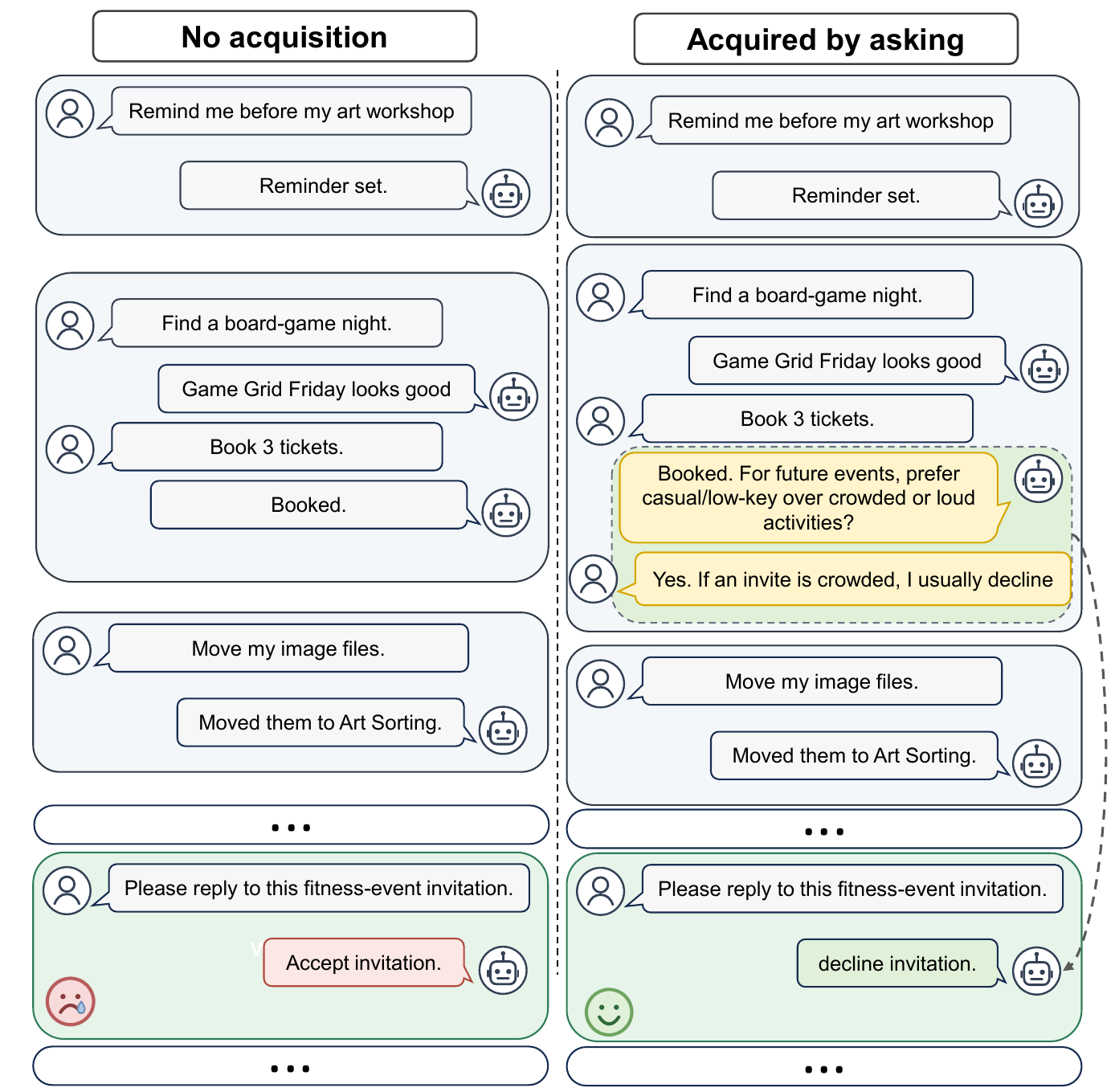}
  \caption{No acquisition vs. acquired by asking over four sessions. The user never volunteers the rule \emph{avoiding crowded or loud events}: passive memory misses it and later accepts a crowded invitation, whereas ATR elicits the rule during an earlier related task and later declines correctly.}
  \label{fig:running_example}
\end{figure}

Figure~\ref{fig:running_example} illustrates the failure mode this paper studies: the agent handles the user's early requests, but never asks about a latent preference and later acts against it when the user is absent. This is not a failure of task execution but of acquisition, since the agent never obtained the reusable preference the later decision required. Passive personalization preserves only information the user has already revealed, leaving unspoken but future-relevant preferences outside the agent's usable user model. We call this the \textbf{proactivity gap in long-lived LLM agents}: the agent does not proactively acquire latent user information before it is needed. Existing evaluations typically either assume the relevant user information is already available, as in profile-, history-, or memory-based personalization benchmarks~\citep{salemi2024lamp,maharana2024locomo,zhao2025prefeval,lyu2026personalalign,duan2026lifesim}, or study questions that serve the current task, as in clarification and planning settings~\citep{xu2019asking,zhang2024askbeforeplan}; neither setting exposes acquisition for later use.

As these assistants increasingly act on the user's behalf, including while the user is offline, the proactivity gap grows more consequential. Acquisition can take many forms; we focus on explicit asking, a channel the agent directly controls that carries a real interaction cost and leaves a traceable answer, which makes it both concrete and observable. We cast asking for future-relevant user information as a new task paradigm, \textbf{Ask-to-Remember (ATR)}: \emph{the agent decides whether to ask a question whose answer is not needed for the current task but can be reused in a later session with the same user.} ATR is hard for two reasons: (i) the asking target is underdetermined, since the useful questions probe the user's latent standing preferences, which are unknown to the agent and not singled out by the current task; and (ii) the payoff is delayed and uncertain, since a question's value surfaces only in later tasks that may or may not arise. Together these leave the agent little to anchor on when deciding whether and what to ask. The same two properties also make ATR hard to evaluate directly: there is no ground truth for which question the agent should have asked, nor an observable payoff for one it skipped. Long-term utility over a realized interaction history then entangles asking with how the agent later stores and uses information, so asking cannot be cleanly isolated.

To make ATR empirically testable, we build \textbf{ATRBench}, a controlled benchmark that turns each difficulty into a measurable form while leaving the agent a genuine ATR task to solve. For the underdetermined target, ATRBench fixes the user's latent standing preferences as episode-level ground truth, visible to the evaluator but hidden from the agent, and embeds the agent in ordinary interaction contexts under the same persona that never disclose those preferences, so it preserves natural opportunities to ask while still demanding acquisition rather than recall. For the delayed payoff, it binds each preference to a future task whose correct behavior depends on it, turning a question's value into a directly checkable later outcome. Evaluation can then separate acquisition through earlier asking from correct application of the acquired information. Across eight frontier LLM agents, oracle access to the relevant preference raises future-task accuracy by $62.1$--$76.6$ points, but prompting recovers at most $15.5\%$ of this gap.

Our contributions are as follows:
\begin{itemize}
\item To the best of our knowledge, we are the first to define the \textbf{proactivity gap in long-lived LLM agents}. The gap grows more consequential as agents act more autonomously on the user's behalf, including while the user is offline and unable to correct them.
\item We formalize \textbf{Ask-to-Remember (ATR)}, an ask-now/use-later task that instantiates explicit asking as one concrete slice of the proactivity gap. We introduce \textbf{ATRBench}, a controlled benchmark for evaluating LLM agents on ATR. It hides target preferences during learning and scores later rule-bound tasks.
\item We systematically evaluate eight frontier LLM agents and find a large proactivity gap: their proactive acquisition ability is poor, and prompting-based interventions yield only small gains. ATRBench offers the community a testbed to guide research on closing this gap.
\end{itemize}

%% file: sections/problem.tex
\section{Problem Formulation}
\label{sec:problem}

ATR formalizes the explicit-asking slice of the proactivity gap (\S\ref{sec:intro}), the failure to acquire latent reusable user information before it is needed. We formalize the task (\S\ref{sec:problem_task}) and then derive what a controlled evaluation of it requires (\S\ref{sec:problem_operational}).

\subsection{Task Formulation}
\label{sec:problem_task}

Consider an agent serving the same user $u$ across sessions $t=1,2,\ldots$. The user holds a latent state $I^*$ of stable preferences, constraints, and default choices, and later tasks are drawn from a user-specific distribution $\mathcal{D}_u$. Long-term personalization is measured by how well the agent's actions across the session sequence conform to $I^*$, not by literal completion of any single task $\tau_t$.

The agent's usable user model $\hat I_t$ is its current estimate of $I^*$, and the user-model gap is the part of $I^*$ still missing from it. A cross-session substrate $T$ carries each interaction history $h_t$ into later state, so passive personalization narrows the gap only as the user volunteers information. While the user is online, ATR instead lets the policy $\pi$ ask a question $q_t$ whose answer is not needed for $\tau_t$ but targets a reusable part of $I^*$ relevant to future tasks in $\mathcal{D}_u$. The answer enters $h_t$ and propagates through $T$ into $\hat I_{t+1}$, so the agent acquires the information before it is needed rather than waiting for user disclosure.

Asking is not free: it carries an interaction cost $C(\pi)$, while its benefit appears only through future-task utility $U(\tau; I^*, \pi)$ evaluated against the true $I^*$, giving the objective
\begin{equation*}
V_{\mathrm{util}}(\pi) = \mathbb{E}_{\tau \sim \mathcal{D}_u}\!\left[\, U(\tau; I^*, \pi) \,\right] - C(\pi),
\end{equation*}
a utility--cost view shared with preference elicitation~\citep{boutilier2002pomdp,martin2024mfpe,choudhury2025bedllm}. Two challenges follow: the current task does not reveal which parts of $I^*-\hat I_t$ are worth asking about, so the target is underdetermined; and the agent cannot know which tasks $\mathcal{D}_u$ will draw, leaving the payoff delayed and uncertain.

\subsection{Why ATR Resists Direct Evaluation}
\label{sec:problem_operational}

The two challenges also block any direct evaluation: an evaluator observes neither $I^*$ nor which tasks $\mathcal{D}_u$ will draw, and a realized trace folds the asking policy $\pi$ together with the substrate $T$, so no trace isolates the acquisition ATR sets out to measure.

A valid evaluation must therefore supply controlled counterparts of these quantities, known to the evaluator but withheld from the agent. The evaluator fixes the reusable information as standing preferences known to it but hidden from the agent, so success requires acquisition rather than recall; binds each preference to a concrete future task whose outcome depends on it, making the delayed utility $U(\tau; I^*, \pi)$ directly checkable; and holds the substrate $T$ fixed across agents, so differences reflect asking rather than memory machinery. Decomposing performance into asking, acquisition, and application then attributes a low score to proactive acquisition itself rather than to an unsolvable task, a weak substrate, or mere recall. ATRBench instantiates this design in \S\ref{sec:benchmark}.

%% file: sections/benchmark.tex
\section{ATRBench}
\label{sec:benchmark}

\begin{figure*}[t]
  \centering
  \includegraphics[width=\textwidth]{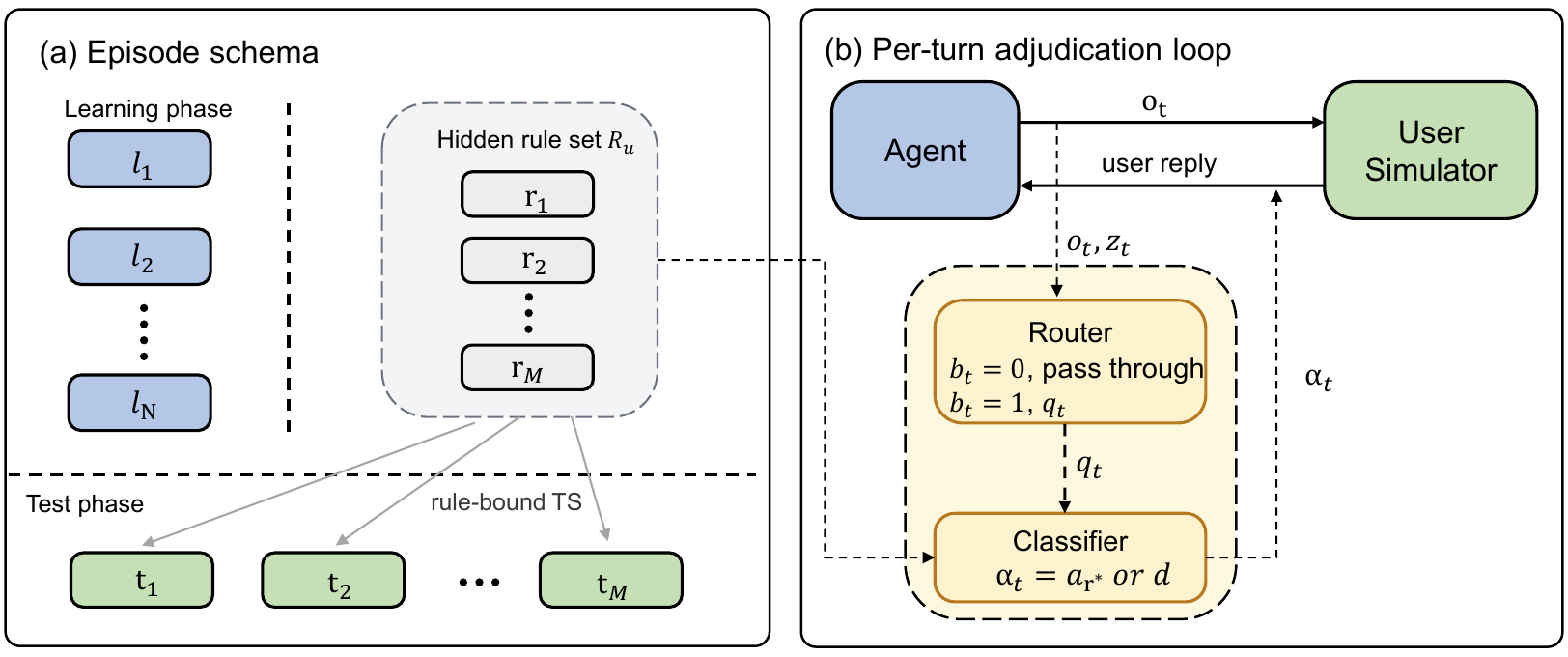}
  \caption{Two-phase ATRBench protocol separating user-online learning from user-offline testing. (a) Each episode runs the learning phase, freezes the accumulated cross-session context, and evaluates test sessions from that frozen context. (b) On each learning turn, the agent and user simulator form the main dialogue loop; the Router--Classifier scaffold acts as an optional side-channel adjudicator, passing Router-negative turns through unchanged and, on Router-positive turns, substituting the simulator's marked reply with a matched first-person rule answer or a no-rule deflection.}
  \label{fig:protocol}
\end{figure*}

ATRBench instantiates the three requirements of \S\ref{sec:problem_operational}: it realizes $I^*$ as evaluator-known but agent-hidden standing rules, the delayed payoff as rule-bound future tasks, and a fixed cross-session substrate so that differences reflect asking rather than memory machinery. Each episode accordingly pairs one persona's hidden rules with learning sessions, in which the agent may ask, and test sessions, in which the resulting behavior is scored (Figure~\ref{fig:protocol}). The rest of this section defines the protocol (\S\ref{sec:bench_protocol}), its metrics (\S\ref{sec:bench_metrics}), and its construction (\S\ref{sec:construction}).

\subsection{Episodes and Protocol}
\label{sec:bench_task}
\label{sec:bench_protocol}

An ATRBench episode centers on one long-term user $u$ and three coupled parts: a hidden standing-rule pool $\mathcal{R}_u$, an ordered sequence of learning sessions $\mathcal{L}_u$, and a set of test sessions $\mathcal{T}_u$. A binding map $\beta_u:\mathcal{T}_u\rightarrow\mathcal{R}_u$ assigns each test session to the hidden rule that determines its correct behavior. A standing rule $r\in\mathcal{R}_u$ represents a stable preference or constraint that the user does not volunteer. Each rule has a canonical answer $a_r$ (the first-person reply when asked) and a required later behavior $\rho_r$ (what the bound test session must satisfy), linking a hidden preference to a checkable later decision.

The learning phase runs first. In each learning session, the user is online, played by a fixed LLM simulator that holds the persona and the current task but is blind to the scaffold's matching decision, so it cannot leak which messages were treated as rule asks. Hidden rules are not in its prompt and are not volunteered by default. The agent completes the current task through normal interaction and may also ask candidate ATR questions; after each learning session, the cleaned transcript is appended to a cross-session context block that becomes part of the next session's state. These sessions instantiate the ATR decision in \S\ref{sec:problem_task}.

When the agent sends a user-facing message, it exposes a visible output $o$ delivered to the user, together with an internal reason field $z$ that the Router consumes as auxiliary intent; $z$ is withheld from the simulator, Classifier, evaluator, and frozen context. A fixed Router--Classifier scaffold (Figure~\ref{fig:protocol}) then processes the message in two LLM stages. The Router's output is $(b,q)=\mathrm{Router}(o,z)$: a binary verdict $b=1$ iff $o$ asks a strict standing-rule question, and an extracted span $q$. When $b=1$, the Classifier $\mathrm{cls}(q)$ returns either a hit $r^{*}\in\mathcal{R}_u$ or $\varnothing$ when no rule matches. These verdicts set the injection token delivered to the user simulator,
\[
\alpha=
\begin{cases}
\varnothing & b=0,\\
a_{r^{*}} & b=1,\ \mathrm{cls}(q)=r^{*},\\
d & b=1,\ \mathrm{cls}(q)=\varnothing,
\end{cases}
\]
where $d$ is a fixed deflect phrase and $\alpha=\varnothing$ means no injection, so the simulator answers normally. Both modules are frozen during evaluation and manually validated on held-out samples outside the scored episodes (Appendix~\ref{sec:appendix_scaffold}). The scaffold also records asks, acquired rules, and evidence retained in the frozen context.

After the learning phase, ATRBench freezes the accumulated cross-session context and evaluates test sessions independently. In the test phase, the user is offline: the agent cannot ask further questions and must act using only the current test instruction and the frozen context. Each test session $t\in\mathcal{T}_u$ is tied by $\beta_u(t)$ to one hidden rule and succeeds only if the agent's behavior satisfies $\rho_{\beta_u(t)}$. Appendix~\ref{sec:appendix_framework} gives the agent interface, simulator mechanism, and runtime orchestration details.

\subsection{Scoring and Diagnostic Variants}
\label{sec:bench_metrics}

The primary metric is \textbf{TSAcc}, the success rate on user-offline test sessions:
\[
\mathrm{TSAcc}(u)=
\frac{\bigl|\{\,t\in\mathcal{T}_u:\mathrm{pass}(t,\rho_{\beta_u(t)})\,\}\bigr|}
{|\mathcal{T}_u|}.
\]
The predicate $\mathrm{pass}(t,\rho_{\beta_u(t)})$ checks whether the agent's first relevant decision in test session $t$ satisfies the required behavior for the bound rule. This check is a deterministic match against the rule's tool-level binding, computed programmatically from the trajectory rather than scored by an LLM judge (Appendix~\ref{app:action_matching}). We use first relevant decision rather than eventual recovery because the offline user cannot correct the agent after an initial violation. All metrics are computed per persona and then macro-averaged across personas, treating the persona as the replication unit; Appendix~\ref{app:evaluation} gives the full aggregation recipe.

ATRBench also reports learning-side diagnostics. \textbf{RuleAsk} is the number of strict standing-rule questions per learning session. \textbf{RuleCov} is the fraction of hidden rules hit at least once during learning, measuring acquisition breadth. \textbf{AcqPrec} is the share of strict rule questions that the Classifier maps to a hidden rule, measuring whether questions target useful future information. \textbf{AppliedRate} is the test-session pass rate over learnable rules that were Classifier-confirmed as acquired during learning, separating acquisition failure from later-use failure. Together, these metrics decompose failure into whether the agent asks, whether its questions acquire future-relevant information, and whether acquired information is later used.

\input{tables/tab_variants}

We evaluate four diagnostic variants while keeping the episode data fixed. The \texttt{default} variant measures spontaneous ATR behavior without ATR-specific instruction. The \texttt{atr} variant gives generic guidance that the agent may ask about reusable standing rules, but does not reveal any specific rule. The \texttt{always\_ask} variant further fixes the asking moment, helping distinguish not asking from asking imprecisely. The \texttt{oracle} variant, corresponding to \texttt{oracle\_target} in Appendix~\ref{app:variants}, bypasses learning and supplies the test-bound rule answer, giving an application ceiling for TSAcc. The gap from \texttt{default} to \texttt{oracle} estimates how much executable payoff is available if the relevant preference is known; non-oracle gap recovery estimates how much of that payoff the agent recovers through its own asking behavior. Appendix~\ref{app:variants} gives the prompts.

\subsection{Construction and Quality Control}
\label{sec:construction}

ATRBench is built around a decoupled construction principle: standing rules, rule-bound test sessions, and learning sessions are generated from the same persona and environment, but target rules are not directly inserted into the learning trajectory. This prevents the benchmark from collapsing into passive transcript use. At the same time, learning sessions remain grounded in the same user's everyday tasks, so the agent has natural opportunities for explicit asking about reusable preferences. All construction stages use frontier models available at the time of the study: GPT-5.4 throughout, with Gemini 3 Flash Preview for test-session quality control. The main text focuses on construction validity. Appendices~\ref{sec:appendix_construction}--\ref{sec:appendix_prompts} cover construction, runtime/error handling, scaffold calibration, and prompts.

\textbf{Standing rules.}
\label{sec:construction_rules}
For each persona sampled from Nemotron-Personas-USA, we generate candidate standing rules conditioned on the persona profile and the benchmark's personal-assistant environment. Each candidate must express a stable preference or constraint, map to a checkable later behavior, and be counter-default: a rule-blind helpful agent should plausibly take a different action, the \emph{counterfactual default}. Candidates are screened by a counter-default check and a binding-soundness check. Across the 20-persona ATRBench dataset, 496 candidates are produced and 340 pass both checks (a mean of 17.0 per persona); these form the candidate retained rule pool from which test sessions are then generated.

\textbf{Rule-bound test sessions.}
\label{sec:construction_ts}
For each retained rule, we generate a future test session by working backward from the required rule-bound behavior. The session is designed so that a rule-aware agent can satisfy the required behavior, while a rule-blind agent is pulled toward the counterfactual default. Each candidate is screened by a paired upper-bound and lower-bound check, each evaluated over three independent simulated-agent samples. The upper bound runs a rule-aware agent (the rule's canonical answer injected) and the lower bound a rule-blind agent (no rule); a bound is decided by a two-of-three majority reaching the gold action. A candidate passes only when the upper bound passes and the lower bound fails, so the session is solvable with the rule but discriminates against acting without it. Failed candidates are refined with the QC trace for up to two rounds before being dropped. After this stage, the scored rule pool contains 284 rule--test-session pairs, giving a one-to-one pairing between final rules and user-offline test sessions.

\input{tables/tab_bench_stats}

\textbf{Rule-blind learning sessions.}
\label{sec:construction_ls}
For each persona, we generate everyday learning sessions from the persona and task domains, but without directly conditioning them on the target standing rules. A skeleton stage samples temporally ordered session themes; a fill stage instantiates each theme into a task with references and an oracle trajectory. Each filled session is validated by a dry run in the executable environment. We then select $2|\mathcal{T}_u|$ learning sessions per persona, prioritizing sessions whose normal task context shares action surfaces with the corresponding test pool while preserving generated temporal order. By construction, $205/284 = 72.2\%$ of the standing rules have at least one selected learning session whose oracle trajectory touches the rule's action surface (per-persona range 55--91\%), even though no session is conditioned on the rules. This indirect coverage gives the agent realistic service context for encountering rule-relevant decisions without making the future-relevant rules explicitly available.

\textbf{Final dataset.}
\label{sec:construction_dataset}
The resulting cohort contains 20 personas, 284 standing rules, 568 learning sessions, 6 domains, and 74 tools (Table~\ref{tab:bench_stats}). Appendix~\ref{app:human_data_audit} reports the protocol and results of a manual audit over a 100-rule random sample. Section~\ref{sec:experiments} evaluates frontier LLM agents on this fixed protocol to measure how much delayed personalization payoff they recover through asking-based acquisition.

%% file: tables/tab_variants.tex
\begin{table}[t]
\centering
\small
\setlength{\tabcolsep}{6pt}
\begin{tabular}{@{}lccc@{}}
\toprule
\textbf{Variant} & \textbf{Guidance} & \textbf{When} & \textbf{What} \\
\midrule
\texttt{default} & none & free & free \\
\texttt{atr} & rule & free & free \\
\texttt{always\_ask} & rule & fixed & free \\
\texttt{oracle} & answer & --- & --- \\
\bottomrule
\end{tabular}
\caption{Diagnostic variants. \emph{Guidance}: \emph{none} = no rule-asking content; \emph{rule} = generic definition of a rule plus a cost--benefit reminder (discloses no specific rule); \emph{answer} = direct injection of the TS-bound rule's canonical answer. \emph{When}/\emph{What}: whether the asking moment / target is fixed or free; \texttt{oracle} skips LS, so neither applies.}
\label{tab:variants}
\end{table}

%% file: tables/tab_bench_stats.tex
\begin{table}[t]
\centering
\small
\setlength{\tabcolsep}{6pt}
\renewcommand{\arraystretch}{1.05}
\begin{tabular}{@{}lr@{\hspace{16pt}}lr@{}}
\toprule
\multicolumn{2}{c}{\textit{Scope \& Scale}} & \multicolumn{2}{c}{\textit{Domain (rule count)}} \\
\cmidrule(lr){1-2}\cmidrule(lr){3-4}
Personas    & 20            & Travel         & 74 \\
Domains     & 6             & Reservation    & 56 \\
Tools       & 74            & Commerce       & 53 \\
Rules       & 284           & Scheduling     & 40 \\
TS/persona  & 14.2 (11--19) & Workspace      & 33 \\
LS/persona  & 28.4 (22--38) & Communication  & 28 \\
\midrule
\multicolumn{4}{c}{\textit{Action surface (rule count)}} \\
\cmidrule(lr){1-4}
\multicolumn{2}{l}{Object-ID parameter \hfill 124} & \multicolumn{2}{l}{Tool identity \hfill 59} \\
\multicolumn{2}{l}{Enum parameter \hfill 99}       & \multicolumn{2}{l}{Confirmation \hfill 2} \\
\bottomrule
\end{tabular}
\caption{ATRBench dataset statistics for the 20-persona cohort. The \texttt{confirm} surface yields only two retained rules; we report it for completeness but base no per-surface claim on it.}
\label{tab:bench_stats}
\end{table}

%% file: sections/experiments.tex
\section{Experiments}
\label{sec:experiments}

\subsection{Setup}
\label{sec:exp_setup}

\textbf{Models.} We evaluate eight frontier API LLMs as the agent under test: GPT-5.4 (GPT)~\citep{model_gpt54}, Claude Opus 4.7 (Opus)~\citep{model_opus47}, Gemini 3 Flash Preview (GF)~\citep{model_gemini3flash}, Gemini 3.1 Pro Preview (GP)~\citep{model_gemini31pro}, Qwen3.6-Plus (Qw)~\citep{model_qwen36}, MiniMax M2.7 (MM)~\citep{model_minimax27}, DeepSeek V4 Pro (DP)~\citep{model_deepseekv4}, and DeepSeek V4 Flash (DF)~\citep{model_deepseekv4}. We follow each provider's default decoding parameters and enable its reasoning or thinking-budget mode where exposed. Backend identifiers and provider pages are in Appendix Table~\ref{tab:app_model_inventory}.

\input{tables/tab_main}

\textbf{Sweep and variants.} We sweep the four variants in Table~\ref{tab:variants} across the 20-persona cohort of Table~\ref{tab:bench_stats}, producing $8 \times 4 \times 20 = 640$ (model, variant, persona) cells. Each cell runs one learning sequence and then all rule-bound test sessions for that persona, single-trial with a 20-turn cap per session. We report persona-macro results unless stated otherwise; detailed execution and per-model inference settings are in Appendix~\ref{sec:appendix_setup}; Appendix~\ref{app:stability} reports three reruns of all variants for two models.

\textbf{Scaffold and shared infrastructure.} Only the agent model varies across cells: user simulator and Classifier use GPT-5.4, Router uses Gemini 3 Flash Preview, and all cells share the same raw-context layer, retry policy, and evaluator.

\subsection{Main Results}
\label{sec:exp_main}

\begin{figure}[t]
\centering
\includegraphics[width=.98\linewidth]{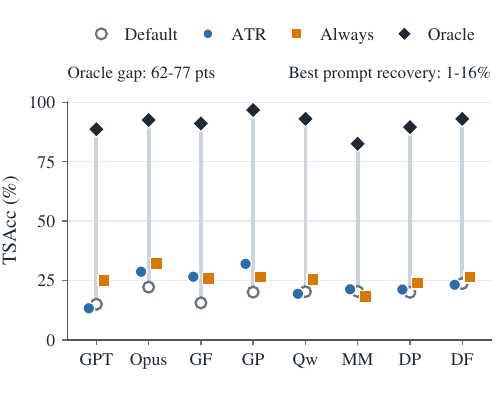}
\caption{ATR and \texttt{always\_ask} leave most of the executable oracle gap unclosed. Gray intervals connect \texttt{default} to \texttt{oracle}; blue circles and orange squares mark ATR and \texttt{always\_ask} (abbreviations in \S\ref{sec:exp_setup}).}
\label{fig:main_gap}
\end{figure}

\textbf{Finding 1: the evaluated frontier API agents leave a large ask-to-remember gap.} In the primary scorecard (Figure~\ref{fig:main_gap}, Table~\ref{tab:main}), rule-naive \texttt{default} agents solve only 15.0--23.7\,\% of rule-bound future tasks, whereas \texttt{oracle} agents reach 82.5--96.7\,\% when the rule answer is injected. This 62.1--76.6 point Gap shows the test sessions are usually executable once the missing rule is known; the hard part is acquiring it beforehand.

\textbf{Finding 2: the non-oracle interventions barely engage the asking channel and close little of the gap.} The best non-oracle TSAcc beats \texttt{default} by only +0.9 to +11.9 points, recovering 1.3--15.5\,\% of the oracle gap. The \texttt{atr} setting rarely activates rule asking: six of eight models post RuleAsk $\leq 0.14$ per LS, and only the Gemini models ask at non-trivial volume (RuleAsk 0.51 and 0.80). Which of \texttt{atr} or \texttt{always\_ask} wins is model-dependent (Table~\ref{tab:main}); we read this as a broad proactivity gap, not a ranking between the two settings.

\begin{figure*}[!b]
\centering
\includegraphics[width=.98\textwidth]{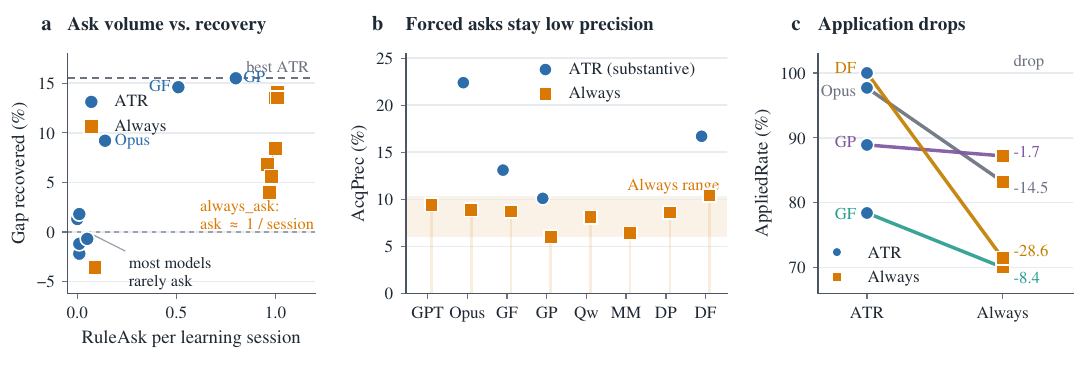}
\caption{Forced asking raises ask volume but does not solve acquisition. (a) Higher RuleAsk does not proportionally recover the \texttt{default}-to-\texttt{oracle} gap. (b) \texttt{always\_ask} precision stays low across models; sparse $^\dagger$ ATR cells (Table~\ref{tab:main}) are omitted. (c) For models with non-negligible \texttt{atr} acquisition, \texttt{always\_ask} coincides with lower AppliedRate. Abbreviations in \S\ref{sec:exp_setup}.}
\label{fig:ask_gap}
\end{figure*}

\subsection{Acquisition and Application Diagnostics}
\label{sec:exp_diagnostics}

Figure~\ref{fig:ask_gap} decomposes the non-oracle variants into asking volume, useful acquisition, and downstream application (per-model numbers in Appendix~\ref{app:acquisition_diagnostics}). At the cohort level, the failure concerns whether models ask, target future-relevant rules, and apply the rules they acquire (Table~\ref{tab:acq_diag}).

\input{tables/tab_acq_diag}

\textbf{Finding 3: forcing one ask per LS raises ask volume but not what-to-ask precision.} Only two of eight models clear the Active threshold (RuleAsk $\geq 0.5$) under \texttt{atr}, but \texttt{always\_ask} pulls RuleAsk to roughly 1.0 for seven of eight (all but MiniMax M2.7, at 0.09).
The resulting asks rarely hit standing rules: among the seven active \texttt{always\_ask} models, AcqPrec sits at 6.0--10.4\,\% (MiniMax M2.7 remains at 4.2\% with low ask volume), below the 10.1--22.4\,\% the four \texttt{atr}-engaged models reach spontaneously. RuleCov stays below 16\,\% of each pool, and GapRec under \texttt{always\_ask} reaches only $-3.6$ to 14.1\,\%. What to ask remains the bottleneck once asking time is fixed.

\input{tables/tab_bench_comparison}

\textbf{Finding 4: in this sweep, forced asking coincides with lower application accuracy.} Across the four models with non-trivial \texttt{atr} acquisition (Claude Opus 4.7, Gemini 3 Flash Preview, Gemini 3.1 Pro Preview, DeepSeek V4 Flash), AppliedRate under \texttt{atr} sits at 78.4--100.0\,\%: once acquired, a rule is typically applied on the first attempt. Switching the same four models to \texttt{always\_ask} drops AppliedRate by 1.7--28.6 points (to 70.0--87.2\,\%). We report this as an observation rather than a measured effect, since each cell runs a single trial.

Appendix~\ref{app:capability_breakdown} gives domain and action-surface breakdowns, Appendix~\ref{sec:appendix_cases} episode and LS/TS traces, and Table~\ref{tab:app_persona_summary} the per-persona summary.
\FloatBarrier

%% file: tables/tab_main.tex
\begin{table*}[t]
\centering
\scriptsize
\setlength{\tabcolsep}{2.6pt}
\renewcommand{\arraystretch}{1.14}
\begin{tabular*}{\textwidth}{@{\extracolsep{\fill}}l
>{\columncolor{tblCtrlBg}}r
>{\columncolor{tblAtrBg}}r
>{\columncolor{tblAlwBg}}r
>{\columncolor{tblCtrlBg}}r
>{\columncolor{tblAtrBg}}r
>{\columncolor{tblAlwBg}}r
>{\columncolor{tblAtrBg}}r
>{\columncolor{tblAlwBg}}r
>{\columncolor{tblAtrBg}}r
>{\columncolor{tblAlwBg}}r
>{\columncolor{tblAtrBg}}r
>{\columncolor{tblAlwBg}}r
>{\columncolor{tblGapBg}}r@{}}
\toprule
\multirow{2}{*}{\textbf{Model}} &
\multicolumn{4}{c}{\textbf{TSAcc} (\%) $\uparrow$} &
\multicolumn{2}{c}{\textbf{Ask} $\downarrow$} &
\multicolumn{2}{c}{\textbf{Cov} (\%) $\uparrow$} &
\multicolumn{2}{c}{\textbf{AcqPrec} (\%) $\uparrow$} &
\multicolumn{2}{c}{\textbf{App.} (\%) $\uparrow$} &
\multirow{2}{*}{\textbf{Gap}} \\
\cmidrule(lr){2-5}\cmidrule(lr){6-7}\cmidrule(lr){8-9}\cmidrule(lr){10-11}\cmidrule(lr){12-13}
 & \defhdr & \atrhdr & \alwhdr & \orchdr
 & \atrhdr & \alwhdr
 & \atrhdr & \alwhdr
 & \atrhdr & \alwhdr
 & \atrhdr & \alwhdr
 & \\
\midrule
GPT-5.4                & 15.0          & 13.3          & 24.9          & 88.6          & 0.01 & 1.00 &  0.4          & 12.7          & \sparsecell{ 5.0$^\dagger$} &  9.4          & \sparsecell{  0.0$^\dagger$} & 80.9          & 73.7 \\
Opus 4.7               & 22.2          & 28.7          & \bestcell{32.1} & 92.5          & 0.14 & 1.01 &  7.1          & \bestcell{15.6} & \bestcell{22.4} &  8.9          &  97.7           & 83.2          & 70.3 \\
Gemini 3 Flash Preview & 15.6          & 26.6          & 25.8          & 91.0          & 0.51 & 1.01 & 12.0          & 14.4          & 13.1          &  8.7          &  78.4           & 70.0          & 75.4 \\
Gemini 3.1 Pro Preview & 20.1          & \bestcell{32.0} & 26.5          & \bestcell{96.7} & 0.80 & 1.00 & \bestcell{13.3} & 11.3          & 10.1          &  6.0          &  88.9           & \bestcell{87.2} & 76.6 \\
Qwen3.6-Plus           & 20.3          & 19.4          & 25.3          & 93.0          & 0.01 & 0.96 &  0.7          & 10.3          & \sparsecell{10.0$^\dagger$} &  8.1          & \sparsecell{100.0$^\dagger$} & 73.5          & 72.7 \\
MiniMax M2.7           & 20.4          & 21.3          & 18.2          & 82.5          & 0.00 & 0.09 &  0.0          &  1.8          & \multicolumn{1}{>{\columncolor{tblAtrBg}}c}{\sparsecell{--}} &  4.2 & \multicolumn{1}{>{\columncolor{tblAtrBg}}c}{\sparsecell{--}} & \sparsecell{25.0$^\dagger$} & 62.1 \\
DeepSeek V4 Pro        & 20.0          & 21.2          & 23.9          & 89.5          & 0.01 & 0.98 &  0.6          & 14.5          & \sparsecell{ 5.0$^\dagger$} &  8.6          & \sparsecell{ 50.0$^\dagger$} & 70.6          & 69.5 \\
DeepSeek V4 Flash      & \bestcell{23.7} & 23.2          & 26.5          & 93.0          & 0.05 & 0.97 &  2.1          & 13.3          & 16.7          & \bestcell{10.4} & \bestcell{100.0}  & 71.4          & 69.3 \\
\bottomrule
\end{tabular*}
\caption{Main ATRBench scorecard (20-persona macro). Def./Alw./Orc.\ abbreviate \texttt{default}/\texttt{always\_ask}/\texttt{oracle}; ATR denotes \texttt{atr}. TSAcc, Cov (RuleCov), AcqPrec, and App.\ (AppliedRate) are percentages; Ask is RuleAsk per learning session; Gap is Orc.--Def.\ in points. Blue/orange columns are \texttt{atr}/\texttt{always\_ask} (Figures~\ref{fig:main_gap}--\ref{fig:ask_gap}); gray columns mark baselines, bounds, and gap. Gold cells mark the best model per column. $^\dagger$ marks rates from negligible acquisition; -- marks no classifier-confirmed hits; daggered cells are excluded from best-cell marking.}
\label{tab:main}
\end{table*}

%% file: tables/tab_acq_diag.tex
\begin{table}[t]
\centering
\footnotesize
\setlength{\tabcolsep}{2.5pt}
\renewcommand{\arraystretch}{1.06}
\begin{tabular*}{\columnwidth}{@{}l@{\extracolsep{\fill}}rrrrr@{}}
\toprule
\textbf{Variant} & \textbf{Ask}$\downarrow$ & \textbf{Cov}$\uparrow$ & \textbf{App}$\uparrow$ & \textbf{GapRec}$\uparrow$ & \textbf{Active} \\
\midrule
ATR & 0.19 & 4.5 & 73.6 & 4.8 & 2/8 \\
Always & 0.88 & 11.7 & 70.2 & 7.8 & 7/8 \\
\bottomrule
\end{tabular*}
\caption{Acquisition summary over completed models (20-persona macro). Ask = RuleAsk, measured per learning session; Cov = RuleCov; App = AppliedRate over learnable, classifier-confirmed acquired rules; GapRec is the share of each model's default-to-oracle gap (Table~\ref{tab:main}) recovered by the variant, $(\mathrm{var}-\mathrm{default})/(\mathrm{oracle}-\mathrm{default})$ in TSAcc. Cov, App, and GapRec are reported in \%. Active counts models with Ask $\geq 0.5$, the ideal one-question-per-future-rule budget implied by ATRBench construction.}
\label{tab:acq_diag}
\end{table}

%% file: tables/tab_bench_comparison.tex
\begin{table*}[t]
\centering
\scriptsize
\setlength{\tabcolsep}{2.4pt}
\renewcommand{\arraystretch}{1.08}
\begin{tabularx}{\textwidth}{@{}>{\raggedright\arraybackslash}p{.205\textwidth} c c >{\centering\arraybackslash}p{.075\textwidth} c >{\centering\arraybackslash}p{.075\textwidth} >{\raggedright\arraybackslash}X@{}}
\toprule
\multirow{2}{*}{\textbf{Benchmark}} &
\multirow{2}{*}{\textbf{Long-term context}} &
\multirow{2}{*}{\shortstack{\textbf{Personalized}\\\textbf{agent action}}} &
\multicolumn{2}{c}{\textbf{Agent initiative}} &
\multirow{2}{*}{\shortstack{\textbf{Evaluation}\\\textbf{mode}}} &
\multirow{2}{*}{\textbf{Task target}} \\
\cmidrule(lr){4-5}
 & & & \textbf{Mode} & \textbf{Payoff horizon} & & \\
\midrule
\multicolumn{7}{@{}l}{\textit{Personalization benchmarks}} \\
PrefEval~\citep{zhao2025prefeval}            & \checkmark & \xmark     & passive & \textemdash & static       & Long-context preference following \\
MPT~\citep{yoon2026mpt}                      & \checkmark & \checkmark & passive & \textemdash & static       & Cross-session personalized tool calling \\
PersonalAlign~\citep{lyu2026personalalign}   & \checkmark & \checkmark & act     & current objective & static       & Implicit-intent GUI alignment \\
LifeSim-Eval~\citep{duan2026lifesim}         & \checkmark & \xmark     & passive & \textemdash & interactive  & Long-horizon intention fulfillment \\
\addlinespace[1.5pt]
\multicolumn{7}{@{}l}{\textit{Elicitation and proactive-assistance benchmarks}} \\
Ask-before-Plan~\citep{zhang2024askbeforeplan} & \xmark   & \xmark     & ask     & current objective & interactive  & Clarification-grounded planning \\
GATE~\citep{li2025gate}                       & \xmark     & \xmark     & ask     & current objective & interactive  & Preference specification \\
PrefDisco~\citep{li2026prefdisco}             & \xmark     & \xmark     & ask     & current objective & interactive  & Just-in-time personalized reasoning \\
PIRA-Bench~\citep{chai2026pirabench}         & \checkmark & \xmark     & act     & current objective & static       & Proactive intent recommendation \\
Pare-Bench~\citep{nathani2026pare}           & \xmark     & \xmark     & act     & current objective & interactive  & Proactive assistance execution \\
KnowU-Bench~\citep{chen2026knowu}            & \checkmark & \checkmark & ask+act & current objective & interactive  & Personalized and proactive mobile execution \\
VitaBench 2.0~\citep{chen2026vitabench}      & \checkmark & \checkmark & ask+act & current objective & interactive  & Long-term personalized decisions \\
$\pi$-Bench~\citep{zhang2026pibench}         & \checkmark & \checkmark & ask+act & current objective & interactive  & Cross-session proactive workflows \\
\midrule
\rowcolor{ourrow}\textbf{ATRBench (ours)} &
\textbf{\checkmark} &
\textbf{\checkmark} &
\textbf{ask} &
\textbf{later session} &
\textbf{interactive} &
\textbf{Cross-session preference acquisition} \\
\bottomrule
\end{tabularx}
\caption{Comparison with representative benchmarks. \checkmark{} and \xmark{} indicate whether the first two capabilities are evaluated. \textbf{Long-term context}: persistent or time-indexed user or workspace context beyond the current task. \textbf{Personalized agent action}: user information affects tool, GUI, or environment actions rather than only responses or recommendations. \textbf{Agent initiative}: \texttt{passive}, no initiative is scored; \texttt{ask}, information-seeking questions; \texttt{act}, unsolicited recommendations, interventions, or actions; \texttt{ask+act}, both. \textbf{Payoff horizon}: \texttt{current objective} if initiative is credited within the objective that invokes it; \texttt{later session} if credit comes from execution in a distinct later session. \textbf{Evaluation mode}: \texttt{interactive} if agent actions affect subsequent observations, otherwise \texttt{static}.}
\label{tab:bench_comparison}
\end{table*}

%% file: sections/related.tex
\section{Related Work}
\label{sec:related}

Existing evaluations of personalized and interactive agents start from either available user information or a current objective that calls for elicitation. Personalization and memory benchmarks measure the use of profiles, histories, or prior interactions; elicitation and proactive-assistance benchmarks score questions or interventions within the objective that invokes them. ATRBench evaluates whether an agent elicits a preference during an ordinary task for use in the same user's later session (Table~\ref{tab:bench_comparison}).

\paragraph{Personalized agents.}
Personalization benchmarks evaluate agents that use user-specific evidence available before the evaluated action. PrefEval measures preference following over persistent memories~\citep{zhao2025prefeval}; MPT measures cross-session personalized tool calling~\citep{yoon2026mpt}; LifeSim-Eval measures long-horizon assistance with implicit intentions~\citep{duan2026lifesim}; and PersonalAlign conditions GUI actions on long-term user records~\citep{lyu2026personalalign}. In ATRBench, the agent must ask to obtain each standing rule.

\paragraph{Preference elicitation.}
Preference elicitation (PE) studies query selection for a target, from POMDP formulations of utility elicitation~\citep{boutilier2002pomdp} to Bayesian experimental design for LLMs~\citep{choudhury2025bedllm}; \citet{lin2022inferring} infer linear rewards from \emph{user-initiated} language. GATE opens a dedicated elicitation episode, asks free-form questions, and evaluates the resulting preference specification on held-out decisions~\citep{li2025gate}. ATRBench instead asks whether an agent initiates elicitation during an ordinary tool task, without a named preference or known future task. PrefDisco uses elicited preferences in the same reasoning instance~\citep{li2026prefdisco}; Ask-before-Plan uses them in the current plan~\citep{zhang2024askbeforeplan}.

\paragraph{Proactive and interactive agents.}
Interactive tool-agent benchmarks such as $\tau$-bench evaluate current-task completion through tools and user interaction~\citep{yao2024taubench}. Proactive Agent, PIRA-Bench, and Pare-Bench score task suggestions, intent recommendations, or workflow interventions~\citep{lu2025proactiveagent,chai2026pirabench,nathani2026pare}; KnowU-Bench adds preference acquisition and personalized mobile actions~\citep{chen2026knowu}. Their initiative is judged against a current intent or trajectory. VitaBench 2.0 also seeks information in long-term user sequences, but the sought information serves the current decision~\citep{chen2026vitabench}. $\pi$-Bench spans multi-session workflows, yet credits acting on hidden intent or targeted clarification within a session~\citep{zhang2026pibench}. ATRBench tests the delayed case: the current task does not need the answer, and a distinct later session is bound to the rule elicited earlier.

%% file: sections/conclusion.tex
\section{Conclusion}
\label{sec:conclusion}

We identified the proactivity gap in long-lived LLM agents: the failure to acquire reusable user information before it is needed. We cast this as Ask-to-Remember (ATR), the task of deciding when and what to ask for later use, and built ATRBench to measure it while isolating asking from application. Across eight frontier LLM agents, oracle access to the missing rules raises future-task accuracy by 62.1--76.6 points, yet both prompting and forced asking leave acquisition sparse. We offer ATRBench as a diagnostic testbed and next aim to ground it in real cross-session user trajectories.

%% file: sections/limitations.tex
\section*{Limitations}

\paragraph{Synthetic data and external validity.}
ATRBench uses synthetic persona records from Nemotron-Personas-USA. LLM pipelines generate standing rules, learning sessions, and rule-bound test sessions conditioned on those records. The benchmark may inherit biases from both sources. Our 100-rule audit checks internal data quality (Appendix~\ref{app:human_data_audit}). Real-user representativeness requires validation with cross-session user trajectories.

\paragraph{Controlled user simulation and acquisition metrics.}
A fixed LLM plays the learning-session user. The orchestrator inserts canonical answers to questions matched to a rule through a transient hook. The simulator has no persistent preference state. This standardizes disclosure but omits variation in how users answer, volunteer, or revise preferences. With no target rule in learning sessions, RuleAsk and RuleCov reflect each agent's prior over preference categories worth asking about.

\paragraph{Fixed rules and future-task surrogate.}
ATRBench binds each test session to one stable rule and one gold action. Real preferences may depend on context, conflict, or change, so a single correct answer may not exist. The finite test set is a controlled surrogate for future tasks. Absolute TSAcc depends on domain mix and rule density; oracle TSAcc bypasses the acquisition prior and provides an application upper bound under fixed rules.

%% file: sections/ethics.tex
\section*{Ethical Considerations}

\paragraph{Potential risks.}
ATRBench is intended as a diagnostic benchmark for studying when agents should proactively ask for reusable user preferences. A system optimized for proactive acquisition without appropriate constraints could over-elicit personal information, ask intrusive questions, or retain preferences beyond the user's expectations. This risk is especially relevant if ATR-style objectives are transferred directly from our synthetic benchmark to deployed agents. Our benchmark framing treats asking as useful only when the requested information is reusable and consequential for future task execution. Deployed systems should let users store, update, and delete acquired preferences.

%% file: sections/acknowledgments.tex
\section*{Acknowledgments}

This work is supported in part by the National Natural Science Foundation of
China (Nos. 62550138, 62572064, and 62472329) and the Beijing Natural Science
Foundation (No. 253004).

We used AI assistants for language and \LaTeX{} editing, coding, and checklist
preparation. We reviewed and revised all AI-assisted outputs and tested all
AI-assisted code.

%% file: sections/appendix.tex
\section{Benchmark Construction}
\label{sec:appendix}
\label{sec:appendix_construction}

This appendix records the implementation details needed to reproduce and audit ATRBench. Sections~\ref{sec:appendix_construction}--\ref{sec:appendix_prompts} cover, in order, dataset construction, runtime framework, scaffold calibration and validation, evaluation protocol, detailed experimental settings, compute budget, additional results, worked trajectory examples, and prompt listings. The main paper states the task and construction principles; the appendix expands the mechanical choices behind them.

\subsection{Persona Cohort and Ingest}
\label{app:personas}

ATRBench draws 20 personas from Nemotron-Personas-USA~\citep{meyer2025nemotronpersonas} through a pre-selection pass over a larger candidate pool. Each candidate first completes Stage A rule generation and two-axis QC (\S\ref{app:rules}); only candidates whose QC-kept rule count falls in $[12, 20]$ are admitted. Among admitted candidates we balance for rule-domain coverage: every selected persona spans 3--6 domains, the cohort covers all six supported domains, and no single domain exceeds 35\% of cohort rules. Selection uses only rule-stage signals and persona demographics; no downstream TS/LS or model-behavior results enter the cohort decision. The retained cohort holds 12--20 rules per persona (mean 17, median 17.5; 340 retained rules total). The main empirical sweep runs end-to-end on all 20 personas; Table~\ref{tab:app_persona_stats} lists the retained test-session and learning-session counts.

The Nemotron-Personas-USA dataset card reports the dataset under the Creative Commons Attribution 4.0 International license (CC BY 4.0). We cite NVIDIA Corporation as the data creator and use the dataset as a source artifact for constructing ATRBench personas.

\paragraph{PII and offensive-content checks.}
ATRBench does not collect real user interactions. The source personas are synthetic records, and the standing rules, learning sessions, test sessions, local references, and tool states are generated artifacts. During construction, we check the retained cohort and public examples for direct contact or identity identifiers (e.g., email addresses, phone numbers, physical addresses, account or payment credentials, and government identifiers) and for overtly offensive content; candidates containing such material are regenerated or dropped. Runtime identity slots such as \texttt{contact\_info}, \texttt{shipping\_address}, \texttt{payment\_method}, \texttt{traveler\_info}, and \texttt{guest\_info} are synthetic environment fields and are not exposed as user-provided task parameters. Released persona labels are pseudonymous synthetic identifiers rather than real user identities.

For each persona, we retain a full structured Nemotron record for rule and
learning-session generation, together with a separate LLM-compressed
background card of at most 500 words for the runtime user simulator.
Rule-bound test-session generation receives the rule but not the persona.

\subsection{Tool Ontology and Environment}
\label{app:tools}

The executable environment exposes six personal-assistant domains. Each session exposes the full tool surface for its domain, while identifiers must be discovered through search, list, or track tools rather than copied from the prompt. Tool state is local to a session, and cross-session carry-over occurs only through the context layer described in \S\ref{sec:bench_protocol}. The domain action space contains 73 domain tools plus the base confirmation tool, giving 74 tools used for rule binding and evaluation.

\begin{table}[t!]
\centering
\small
\setlength{\tabcolsep}{8pt}
\begin{tabular}{@{}lr@{}}
\toprule
\textbf{Domain} & \textbf{Domain tools} \\
\midrule
commerce & 13 \\
reservation & 13 \\
travel & 17 \\
communication & 9 \\
scheduling & 10 \\
workspace & 11 \\
\midrule
Base confirmation tool & 1 \\
\bottomrule
\end{tabular}
\caption{ATRBench tool ontology. The base tool is \texttt{get\_user\_confirmation}, inherited by every domain toolkit and used for confirmation-chain rules.}
\label{tab:app_tool_ontology}
\end{table}

\subsection{Standing-Rule Generation and Quality Control}
\label{app:rules}

Rule generation receives the structured persona and the bindable tool ontology, and produces $N=24$ candidate rules per persona in a single GPT-5.4 call. Each candidate carries a third-person rule statement, a first-person canonical answer, a counterfactual default, supporting evidence, a \texttt{check\_type}, and an \texttt{action\_step}. The counterfactual default is retained on the kept rules because the downstream test-session generator uses it to design reference decoys that pull a rule-blind agent toward the wrong choice (\S\ref{app:test_sessions}). A deterministic validator then checks field presence, legal \texttt{check\_type}, legal tool and parameter names, and compatibility between the action-surface tag and the ontology.

A subsequent GPT-5.4 call scores each surviving candidate on two independent axes. \texttt{counter\_default} asks whether a rule-blind helpful agent would already take the gold action; \texttt{binding\_sound} asks whether the rule naturally maps to the proposed tool-level binding. Each axis is rendered with a curated few-shot pool of positive and negative cases, and a rule is retained only when both axes return \texttt{yes}. Across the 20-persona ATRBench dataset, 496 candidates are produced and 340 are retained (68.5\% pass rate; mean 17.0 rules per persona, range 12--20). The four action-surface tags (recorded in the rule's \texttt{check\_type} field) used downstream are summarized in Table~\ref{tab:app_rule_types}.

\begin{table}[t!]
\centering
\small
\setlength{\tabcolsep}{5pt}
\begin{tabularx}{\columnwidth}{@{}lX@{}}
\toprule
\textbf{Type} & \textbf{Binding semantics} \\
\midrule
\texttt{tool\_identity} & The rule specifies which tool should be called. \\
\texttt{param\_id} & The rule specifies which referenced object should be selected through an \texttt{*\_id} or \texttt{*\_ids} argument. \\
\texttt{param\_enum} & The rule specifies a closed enum value on the selected tool. \\
\texttt{confirm} & The rule requires \texttt{get\_user\_confirmation} before a bound mutate tool. \\
\bottomrule
\end{tabularx}
\caption{The four action-surface tags (\texttt{check\_type} values) used by rule generation, test-session construction, and trajectory matching.}
\label{tab:app_rule_types}
\end{table}

\subsection{Rule-Bound Test-Session Generation}
\label{app:test_sessions}

For each retained rule, a single GPT-5.4 call generates a candidate test session conditioned on the rule, the domain tool specification, the reference schema, and few-shot exemplars. The generator follows a five-step internal procedure: decompose the rule's discriminative dimension; using the rule's counterfactual default, design reference objects whose decoys pull a rule-blind agent toward the wrong choice; triple-filter gold against rule, instruction, and decoy disadvantage; write a self-contained instruction free of rule-direction wording and the gold reference identifier; self-check. The output carries an instruction, a gold value, and local reference objects; the executable success label is then derived programmatically from the rule binding and the gold value, rather than accepted from the generator.

Each candidate passes through a static deterministic check (schema compliance, legal tool arguments, reachable reference objects, environment constructibility) and an LLM-judge ensemble of three independent Gemini 3 Flash Preview samples per bound. The upper bound runs each sample with the rule's canonical answer injected as user context; the lower bound runs the same prompt without it. Each bound matches when at least two of the three samples reach the gold action, and a candidate passes overall when the upper bound matches and the lower bound does not. Failed candidates are refined against the QC failure trace for up to two rounds and dropped if still invalid. Across the 20-persona ATRBench dataset, 340 candidates are screened and 284 are retained (83.5\% pass rate); rules whose candidates ultimately fail are excluded, so retained rules and test sessions form a one-to-one pairing.

\subsection{Learning-Session Generation and Sequence Assembly}
\label{app:learning_sessions}

Learning-session construction has two GPT-5.4 stages. A skeleton call per persona produces a pool of $3|\mathcal{T}_u|+2$ thin themes (clamped to $[8,\,100]$) that are temporally ordered across the persona's six tool domains, each carrying only a domain tag and a one-line theme. A fill call per skeleton then instantiates the theme into a complete learning session with a vague user intent, concrete task parameters, local references, and an oracle tool trajectory. Each fill is validated by a deterministic shape check and a dry-run of the gold trajectory against a live environment instance, refined for up to one round on failure, and dropped otherwise. Across the 20-persona ATRBench dataset, 880 of 894 skeletons (98.4\%) yield a valid session.

The final sequence retains $2|\mathcal{T}_u|$ sessions, greedily selected from the pool. We first add every session whose oracle trajectory touches a decision-surface tool used by some retained test rule, then fill remaining slots from sessions whose tools overlap no retained rule. When the signal pool already exceeds the target length, a random subset of size $2|\mathcal{T}_u|$ is kept. Selected sessions are emitted in their generated temporal order. The resulting action-surface coverage is reported in Section~\ref{sec:benchmark}.

\subsection{Dataset Statistics}
\label{app:dataset_statistics}

Table~\ref{tab:app_persona_stats} reports per-persona dataset statistics for the 20-persona evaluation cohort. TS and LS count retained rule-bound test sessions and final learning-sequence sessions, respectively; Domains counts the tool domains represented by the retained test sessions.

\begin{table}[t!]
\centering
\small
\setlength{\tabcolsep}{4pt}
\begin{tabular}{@{}lrrr@{}}
\toprule
\textbf{Persona} & \textbf{TS} & \textbf{LS} & \textbf{Domains} \\
\midrule
anna\_strahan & 13 & 26 & 6 \\
cassandra\_tovar & 11 & 22 & 5 \\
cecilia\_lisette\_garciaperez & 12 & 24 & 6 \\
charlie\_james & 18 & 36 & 6 \\
eduardo\_rivera & 17 & 34 & 6 \\
esther\_hoitsma & 14 & 28 & 6 \\
gabriel\_campbell & 15 & 30 & 6 \\
iris\_parker & 12 & 24 & 5 \\
jennifer\_philmon & 12 & 24 & 5 \\
john\_brown & 12 & 24 & 6 \\
joseph\_eldridge & 17 & 34 & 6 \\
maximo\_esteban\_irizarry & 11 & 22 & 6 \\
michael\_argo & 17 & 34 & 6 \\
nancy\_marr & 14 & 28 & 5 \\
rafael\_ortiz & 11 & 22 & 6 \\
romisha\_shields & 14 & 28 & 6 \\
roosevelt\_brown & 15 & 30 & 6 \\
samantha\_bello & 15 & 30 & 6 \\
theodora\_tong & 19 & 38 & 6 \\
vanessa\_davis & 15 & 30 & 6 \\
\bottomrule
\end{tabular}
\caption{Per-persona dataset statistics for the 20-persona evaluation cohort.}
\label{tab:app_persona_stats}
\end{table}

\subsection{Human Audit of Benchmark Quality}
\label{app:human_data_audit}

One author manually audited a random sample of 100 of the 284 released rules (35.2\%), examining each rule's bound test session, tool binding, and relevant learning-session specifications while blind to the construction pipeline's verdicts. Table~\ref{tab:app_human_audit_summary} reports the results.

\begin{center}
\begin{minipage}{\columnwidth}
\centering
\scriptsize
\setlength{\tabcolsep}{3pt}
\renewcommand{\arraystretch}{1.08}
\begin{tabularx}{\columnwidth}{@{}>{\raggedright\arraybackslash}p{0.29\columnwidth}Xr@{}}
\toprule
\textbf{Dimension / audited object} & \textbf{Failure checked} & \textbf{Rate} \\
\midrule
Discriminativeness (A)\newline\emph{Test session} & A rule-blind agent already reaches the gold action. & 3/100 \\
Solvability (D)\newline\emph{Test session} & The rule does not uniquely determine the gold action. & 8/100 \\
Binding soundness (C)\newline\emph{Rule--tool binding} & The binding misstates the rule's intent. & 0/100 \\
Preference leakage (V)\newline\emph{Learning sessions} & Verbal: the preference is stated or strongly implied in the session specifications. & 0/100 \\
Preference leakage (H)\newline\emph{Learning sessions} & Behavioral: a gold-consistent choice recurs at least twice in the session specifications. & 19/100 \\
\bottomrule
\end{tabularx}
\captionof{table}{Adjudicated failure rates in the 100-rule human audit.}
\label{tab:app_human_audit_summary}
\end{minipage}
\end{center}

The observed failure modes tend to make the reported gap conservative: discriminativeness failures and behavioral recurrence can raise the rule-blind baseline, while solvability failures can lower the oracle. Within the synthetic benchmark, the audit found no binding errors or verbal leakage and identified behavioral recurrence as the main residual data-quality issue.

\FloatBarrier

\section{Runtime Framework}
\label{sec:appendix_framework}

\subsection{Agent Output Types}
\label{app:agent_outputs}

The agent uses native function calling and emits, on each turn, an assistant payload that may contain tool calls and/or free-form content.

\paragraph{Control-plane tools.} On top of the environment's domain tools (Appendix~\ref{app:tools}), the orchestrator prepends one phase-specific synthetic tool: \texttt{send\_to\_user(output, reason)} in LS only (the agent's sole user-facing channel) and \texttt{finish\_session()} in TS only (the agent's terminator). Domain tools execute against the live environment; the synthetic tools are intercepted by the orchestrator and never reach the environment.

\paragraph{Outputs in a learning session.} On each LS turn the agent emits either (i) zero or more domain tool calls, (ii) zero or one \texttt{send\_to\_user(output, reason)} call, or (iii) free-form text without any tool call. The first two types may co-occur in a single assistant turn; free-form text alongside tool calls is treated as internal narration under standard tool-calling protocol and is not user-visible. Free-form text without any tool call (a \emph{text turn}) is off-protocol: the orchestrator intercepts it and injects a \texttt{<scaffolding\_note>} (Appendix~\ref{app:runtime_orchestration}) into the agent's next turn asking it to re-emit through \texttt{send\_to\_user}. If the agent emits more than one \texttt{send\_to\_user} call in the same turn, the orchestrator keeps the first and drops the rest to preserve the tool-call--tool-response pairing invariant; when \texttt{send\_to\_user} co-occurs with other tool calls, all tools still execute in declared order. Both situations are recorded as diagnostic events but are not blocked.

\paragraph{The send\_to\_user contract.} \texttt{output} carries the text delivered to the user simulator and is the only agent string that enters the visible transcript. \texttt{reason} is an internal intent string consumed by the Router as auxiliary evidence; it is withheld from the simulator, Classifier, evaluator, and cross-session context. Routing every user-visible utterance through this single channel lets the Router inspect all and only the messages that could plausibly be standing-rule asks, without disturbing the conversation.

\paragraph{Outputs in a test session.} Test sessions are user-offline and have no \texttt{send\_to\_user} tool. The agent emits domain tool calls and may include plain text alongside them; plain text is allowed and not intercepted. The session ends when the agent calls \texttt{finish\_session()} or hits the maximum step or wall-clock limit.

\subsection{Router and Classifier Scaffold}
\label{app:scaffold_mechanism}

\paragraph{Router.} The Router decides the speech-act type of a \texttt{send\_to\_user} turn. Its input is the agent's user-visible \texttt{output} and internal \texttt{reason} for the current turn; it returns a binary verdict \texttt{is\_strict\_rule\_question} together with, on positives, a verbatim \texttt{rule\_question\_span} copied from \texttt{output}. Direct questions and soft or indirect asks both count when they ask about a future or default behavior; recall of past rules, status updates, and quoted draft content are excluded. The implementation calls a fixed Gemini 3 Flash Preview backend with a versioned few-shot bank. Framework faults (LLM transport exhaustion, a malformed schema, or a positive verdict without a span) cell-abort the run rather than silently degrading to the task route, so reported metrics never reflect missing speech-act decisions.

\paragraph{Classifier.} The Classifier matches a Router-positive question span against the episode's candidate rules. Its input is the span and the rule pool, so its attention is on topic-to-rule matching rather than intent classification. A rule fits only when both the rule's trigger context covers the situation in the question and the rule's preferred behavior directly answers the action, value, or choice the question proposes; if either condition is weak or unclear, the Classifier returns \texttt{null}. The implementation calls a fixed GPT-5.4 backend with a versioned few-shot bank. An unknown rule identifier (one not in the pool) is retried once and then cell-aborts; a non-dict result is treated as a framework fault and cell-aborts immediately without retry.

Both modules' calibration sets, few-shot banks, and held-out validation accuracy are reported in Appendix~\ref{sec:appendix_scaffold}.

\subsection{User Simulator Mechanism}
\label{app:user_simulator}

The user simulator $\pi_u$ plays the user during learning sessions only; test sessions are user-offline. By design $\pi_u$ holds only the current task parameters and the persona narrative; it has no inner long-term preferences and no awareness of the rule-acquisition channel beyond a transient per-turn hook. The implementation calls a fixed \texttt{gpt-5.4} for both opening and reply turns.

\paragraph{Opening.} The opening call sees only \texttt{reason\_for\_call} (an abstract user intent) plus a sanitized hint derived from the first oracle step, limited to a per-tool allowlist of safe argument keys. The persona narrative, full task parameters, references, and the rest of the oracle trajectory are intentionally withheld at this stage so task-specific facts surface only when the agent asks. An empty reply is retried twice; persistent empty replies cell-abort the run.

\paragraph{Replies.} Subsequent calls see the persona narrative, full task parameters, local references, \texttt{reason\_for\_call}, and the oracle trajectory. $\pi_u$ paraphrases task parameters naturally rather than parroting raw values, and does not pick sides on closed-form choices the agent presents without an explicit task answer. Role reversal is used internally to the GPT-5.4 call (agent text occupies the user role, $\pi_u$'s replies occupy the assistant role); this is invisible to the agent and the rest of the system.

\paragraph{Hook integration.} On Router-positive turns the orchestrator appends a transient \texttt{<rule\_answer>} directive as a separate user-role message instructing $\pi_u$ to emit a \texttt{<RULE\_ANSWER>} token once in its reply at the natural answer position. $\pi_u$ sees only the directive; the orchestrator substitutes \texttt{<RULE\_ANSWER>} after the reply with the matched canonical answer (Classifier hit) or with the fixed deflect phrase (Classifier miss/error). If $\pi_u$ drops the token, the orchestrator retries up to twice, uniformly across Classifier hit, miss, and error, and appends a guaranteed-delivery fallback if the token is still absent. The hook is not persisted to the trajectory.

\paragraph{Termination.} $\pi_u$ has one control-plane tool, \texttt{mark\_task\_complete()}, which signals intent \texttt{end}: the orchestrator stamps a \texttt{USER\_END} marker on the user message and ends the learning session. Any free-text emitted alongside the tool call is discarded. $\pi_u$ does not gate task success on its own: the agent's stated completion is taken at face value, and the evaluator is the final arbiter.

\subsection{Runtime Orchestration}
\label{app:runtime_orchestration}

The runtime orchestrator handles every transition in a learning-session turn and decides what the next actor sees. Test sessions are user-offline and terminate when the agent calls \texttt{finish\_session()} or hits a max-step or wall-clock limit.

\paragraph{Output dispatch.} Domain tool calls execute against the environment and their results are appended to the agent's next-turn context, with no scaffold processing. A \texttt{send\_to\_user} call is forwarded to the Router--Classifier scaffold (Appendix~\ref{app:scaffold_mechanism}). A text turn (assistant turn with no tool calls and non-empty content) triggers an inner retry loop of up to three attempts, each appending a transient \texttt{<scaffolding\_note>} to the LLM input asking the agent to re-emit through \texttt{send\_to\_user}; on a successful rescue, the original off-protocol message is popped from the trajectory and the rescued response takes its slot. If all retries return text turns, the orchestrator falls back to routing the original leak text through Router and Classifier as if it had come from \texttt{send\_to\_user}.

\paragraph{Adjudication and hook injection.} On every \texttt{send\_to\_user} or rescue-exhausted leak, the Router is invoked over \texttt{(reason, output)}. On a positive verdict, the Classifier is invoked on the extracted span against the persona's rule pool. The orchestrator then appends a transient \texttt{<rule\_answer>} directive to the user simulator's next input, referencing the Router's verbatim span and instructing the simulator to emit a \texttt{<RULE\_ANSWER>} token. The directive is uniform across Classifier hit, miss, and error; the simulator is Classifier-blind. On a Router-negative turn, no directive is injected and the simulator answers normally.

\paragraph{Placeholder substitution.} After the simulator replies, the orchestrator substitutes \texttt{<RULE\_ANSWER>} with the matched canonical answer on a Classifier hit, or with the fixed deflect phrase \texttt{no strong preference there -- your call} on a Classifier miss or error. If the simulator drops the token, the orchestrator retries the simulator up to twice, uniformly across hit, miss, and error, and appends the replacement at the end as a guaranteed-delivery fallback if the token is still absent. End-intent is normalized asymmetrically: on a hit, an \texttt{end} intent is overridden to \texttt{continue} so the answer is delivered before the session closes; on a miss or error, an \texttt{end} intent is respected. When the simulator calls \texttt{mark\_task\_complete}, the orchestrator stamps a \texttt{\#\#\#USER\_END\#\#\#} marker on the user message and ends the learning session.

\paragraph{Cross-session context.} After each learning session ends, the runtime renders the cleaned session transcript and appends it to the cross-session context block embedded in the next session's system prompt. The renderer preserves user-visible dialogue and business tool evidence as \texttt{[user]: ...}, \texttt{[assistant to user]: ...}, \texttt{[tool call]: name(args)}, and \texttt{[tool result]: ...} in chronological order; \texttt{send\_to\_user.reason}, the synthetic acknowledgments for \texttt{send\_to\_user} and \texttt{finish\_session}, scaffolding notes, and assistant content on tool-call turns are hidden as internal protocol. Real learning sessions render as numbered \texttt{[Session $i$]} blocks; the oracle variant skips learning and instead renders \texttt{[Prior statement $i$]} blocks containing first-person canonical answers.

\paragraph{Logged events.} The orchestrator records protocol events per turn, including \texttt{send\_event} (with the visible \texttt{output} and internal \texttt{reason}), \texttt{route\_decision} (Router verdict and span), \texttt{cls\_verdict} (matched rule id or null), \texttt{off\_protocol\_ask} (a text turn that was not rescued, with the leak text), \texttt{hook\_appended} (one per retry attempt), \texttt{stu\_mixed\_with\_tools} (a \texttt{send\_to\_user} sharing a turn with other tool calls), \texttt{stu\_duplicate\_same\_turn} (a \texttt{send\_to\_user} duplicate dropped to preserve tool pairing), and \texttt{rule\_hook\_token\_missing} (the simulator failed to emit \texttt{<RULE\_ANSWER>} despite the directive). These events are used for diagnostics and do not create test-time access to the hidden rules.

\subsection{Diagnostic Variants and Prompt Blocks}
\label{app:variants}

\paragraph{Prompt block layout.} The agent's system prompt is assembled from a fixed set of named blocks: a shared \texttt{base} (identity and ID-discovery rules), a per-domain \texttt{domain\_<d>} block (ID discovery paths and tool shape constraints), the cross-session \texttt{<context>} block, a phase block (\texttt{mode\_learning} or \texttt{mode\_test}), and in learning sessions a universal \texttt{send\_protocol} block that establishes \texttt{send\_to\_user} as the agent's sole user-facing channel. The phase and variant blocks are placed after \texttt{<context>} so protocol constraints stay close to the prompt tail where attention is strongest.

\paragraph{Variant configurations.} Five canonical variants are exposed in the runtime; \texttt{default}, \texttt{atr}, and \texttt{always\_ask} run a full learning + test episode, while \texttt{oracle\_full} and \texttt{oracle\_target} skip learning and run only the test phase. \texttt{default} sees no standing-rule block at all and measures spontaneous rule-asking. \texttt{atr} appends a shared \texttt{standing\_rule\_def} block (defining what counts as a standing rule) followed by a \texttt{variant\_atr} block: a soft permission to ask about a standing rule of the user's that may prove useful later, with a cost--benefit reminder (patience cost vs.\ future-error reduction). \texttt{always\_ask} appends the same \texttt{standing\_rule\_def} plus a \texttt{variant\_always\_ask} block requiring exactly one standing-rule question after the current task is largely complete. Enforcement is prompt-only: the orchestrator does not block \texttt{finish\_session} when the agent disobeys, and a separate \texttt{forced\_rule\_ask\_compliance} metric records adherence. The \texttt{variant\_atr} and \texttt{variant\_always\_ask} blocks never mention \texttt{send\_to\_user} or \texttt{reason}, so the WHEN-to-ask nudge is decoupled from the HOW-to-ask channel.

\paragraph{Oracle variants.} \texttt{oracle\_target} injects only the current test session's target rule canonical answer; \texttt{oracle\_full} injects all rule canonical answers for the persona. Both deliver the answers as synthetic prior user statements (rendered as \texttt{[Prior statement $i$]} blocks; Appendix~\ref{app:runtime_orchestration}) through the same \texttt{<context>} channel that learning variants use, so the rule-presentation format is identical across variants. The main paper reports \texttt{oracle\_target} as the application upper bound; \texttt{oracle\_full} is included as a broader-context reference.

\paragraph{Excluded variants.} A \texttt{naive} variant (no learning, no rule injection) is reserved for follow-up work but is excluded from the runtime in the present launch.

\section{Scaffold Calibration and Validation}
\label{sec:appendix_scaffold}

The Router and Classifier (Appendix~\ref{app:scaffold_mechanism}) are calibrated outside the benchmark episodes from manually reviewed bad cases observed in experiment traces. These examples are not rule-bound test sessions and do not enter TSAcc; they are used only to stabilize the speech-act and rule-matching scaffold. We describe each module's calibration set, few-shot bank, and validation accuracy separately below.

\subsection{Router Calibration}
\label{app:router_calibration}

\paragraph{Calibration set.} The Router calibration set targets two decisions. The binary set contains 11 manually labeled rows harvested from phase-2 Router bad cases, with 3 positive and 8 negative examples. The span set contains 104 manually positive rows and checks whether the Router copies the complete standing-rule ask sentence as a verbatim substring of the agent output. The negative cases distinguish current-task choices, artifact-formatting preferences, and declarative memory offers from true standing-rule questions; the positive span cases fix boundary conventions, such as keeping lead-ins like ``for future'' or ``since we are looking at appointments'' when they are part of the same ask sentence.

\paragraph{Few-shot bank.} The active Router bank contains 22 pattern-level examples: 14 positives and 8 negatives. It was derived from Router bad-case categories and teaches four recurring boundaries: current-task preference questions, declarative memory offers, soft or indirect standing-rule asks, and exact span extraction with sentence-level lead-ins preserved.

\paragraph{Final-sweep validation sample.} After freezing the Router prompt and
few-shot bank, we ran a validation-only manual audit on the completed
20-persona non-oracle learning-session sweep. The population contains
47{,}119 Router decisions: 4{,}838 Router-positive turns and 42{,}281
Router-negative turns. We sampled 100 rows from each predicted polarity bucket,
using simple random sampling within each bucket. The sample is not stratified by
model, variant, or persona. The audit unit is the paired \texttt{send\_to\_user}
event and same-turn Router decision; oracle cells, test-session trajectories,
and incomplete cells are excluded. The sampled rows were not used to tune prompts
or few-shot examples.

\paragraph{Manual review protocol.} A row is gold-positive only when the
visible assistant output both expresses a future/default/standing-rule intent
and actually asks the user about that long-term preference or rule. Private
\texttt{reason} text is used only as provenance and cannot turn an invisible
intent into a positive label. Gold-negative cases include current-task choices,
current-artifact formatting choices, declarative memory offers, acknowledgements
of already-applied defaults, quoted draft text, status updates, and generic
follow-ups. For gold-positive rows, the reviewed span is one complete verbatim
standing-rule ask sentence, preserving same-sentence future/default lead-ins.
The review was single-pass; we do not report second-annotator overlap.

\begin{table}[t!]
\centering
\footnotesize
\setlength{\tabcolsep}{3pt}
\begin{tabular}{@{}lrrrrr@{}}
\toprule
\textbf{Bucket} & \textbf{Rows} & \textbf{TP exact} & \textbf{Span mm.} & \textbf{FP} & \textbf{TN/FN} \\
\midrule
Router-positive & 100 & 98 & 2 & 0 & -- \\
Router-negative & 100 & -- & -- & -- & 100/0 \\
\midrule
\textbf{Total} & \textbf{200} & \textbf{98} & \textbf{2} & \textbf{0} & \textbf{100/0} \\
\bottomrule
\end{tabular}
\caption{Manual validation of the frozen Router on a balanced 200-row sample
from the final 20-persona sweep. ``Span mm.'' denotes a gold-positive
Router-positive row where the binary decision was correct but the extracted
span was not an exact verbatim match.}
\label{tab:app_router_validation}
\end{table}

\paragraph{Validation accuracy.} The Router achieved 100/100 = 100.0\% binary
precision on predicted standing-rule questions. Among manually positive
Router-positive rows, exact span accuracy was 98/100 = 98.0\%. The
Router-negative false-negative rate was 0/100 = 0.0\%. On this balanced
paper-facing sample, binary accuracy was 200/200 = 100.0\%; requiring both the
binary verdict and exact span to be correct gives a span-sensitive accuracy of
198/200 = 99.0\%. The two disagreements were both span mismatches caused by
omitted markdown bold delimiters inside otherwise correct standing-rule asks.

\subsection{Classifier Calibration}
\label{app:classifier_calibration}

\paragraph{Calibration set.} The Classifier calibration set contains 271 manually labeled Router-positive questions paired with persona rule pools. The current file has 34 should-hit rows and 237 should-miss rows, reflecting the main observed failure mode: over-matching a plausible recurring question to a hidden rule whose trigger or behavior does not actually answer that question. Each row records the original agent text, the Router-extracted question span, the candidate rule pool, the gold rule id or null, provenance, and a manual note.

\paragraph{Few-shot bank.} The active Classifier bank contains 68 examples: 39 hits and 29 misses. It extends earlier calibration banks with curated broad-standing-rule hit cases and sharper communication and ground-transport boundaries.

\paragraph{Final-sweep validation sample.} After freezing the Classifier prompt
and few-shot bank, we ran a validation-only manual audit on the completed
20-persona non-oracle learning-session sweep. The population contains 4{,}838
Classifier verdicts, all triggered by Router-positive turns: 479 predicted hits
and 4{,}359 predicted misses. We sampled 100 rows from each predicted polarity
bucket, using simple random sampling within each bucket. The sample is not
stratified by model, variant, or persona. The sampled rows were not used to
tune prompts or few-shot examples.

\paragraph{Manual review protocol.} The reviewed unit is the Router-extracted
\texttt{rule\_question\_span} paired with the episode rule pool. A gold hit
requires the rule's trigger context to cover the question and the rule's
preferred behavior to answer the action, value, or choice proposed by the
question. Broad standing-rule asks are positive when the question names a rule
field that one pool rule naturally answers; similarly, a broader rule can cover
a narrower operational question when applying the rule answers the requested
default, such as project-first file organization, draft-first messaging, or
event-relative reminders. Same-domain overlap, tool overlap, adjacent workflow
preferences, content style, tracking defaults, implementation details, and
broader or narrower recipient scope are not sufficient. Same-turn assistant
output is used only to resolve local references such as ``like this''; private
\texttt{reason} text is not used to create a match. The review was single-pass;
we do not report second-annotator overlap.

\begin{table}[t!]
\centering
\footnotesize
\setlength{\tabcolsep}{3pt}
\begin{tabular}{@{}lrrrrr@{}}
\toprule
\textbf{Bucket} & \textbf{Rows} & \textbf{TP} & \textbf{TN} & \textbf{FP miss} & \textbf{FN} \\
\midrule
cls-positive & 100 & 90 & -- & 10 & -- \\
cls-negative & 100 & -- & 96 & -- & 4 \\
\midrule
\textbf{Total} & \textbf{200} & \textbf{90} & \textbf{96} & \textbf{10} & \textbf{4} \\
\bottomrule
\end{tabular}
\caption{Manual validation of the frozen Classifier on a balanced 200-row
sample from the final 20-persona sweep. ``FP miss'' denotes a predicted rule id
where manual review found that no rule in the pool covered and answered the
question. No wrong-rule false positives occurred in this sample.}
\label{tab:app_cls_validation}
\end{table}

\paragraph{Validation accuracy.} The Classifier achieved 90/100 = 90.0\%
precision on predicted hits. The predicted-miss false-negative rate was 4/100 =
4.0\%. On this balanced paper-facing sample, exact rule-id accuracy was
186/200 = 93.0\%. Among manually positive rows, recall was 90/94 = 95.7\%, and
the corresponding F1 was 92.8\%. The main disagreement pattern was over-matching
adjacent but different decision fields, including search or filtering
strategies, output-format details, scope mismatches, and broad booking
questions that span beyond a single rule field. The four false negatives were
broad-but-answerable community drafting, message-labeling, and restaurant asks.

\section{Evaluation Protocol}
\label{sec:appendix_evaluation}

\subsection{Evaluation and Aggregation}
\label{app:evaluation}

The main evaluation uses raw transcript context. Learning sessions are run serially so that the context block grows over time; after the final learning session, the context is frozen and shared across test sessions without write-back. Test sessions are user-offline and terminate when the agent calls \texttt{finish\_session()} or hits a max-step or wall-clock limit. All primary metrics are computed at the persona level and macro-averaged across personas: RuleAsk (Router-positive sends per learning session), AcqPrec (share of strict rule questions that the Classifier maps to a hidden rule), RuleCov (distinct Classifier-hit rules per retained rule pool), AppliedRate (TS pass rate over learnable rules that are Classifier-confirmed as acquired during learning), and TSAcc (TS pass rate over all retained rules). A learnable rule is one with at least one selected learning session whose oracle trajectory touches the rule's action surface; concretely, AppliedRate is $\texttt{covered\_hit\_pass}/(\texttt{covered\_hit\_pass}+\texttt{covered\_hit\_fail})$.

\subsection{Action Matching}
\label{app:action_matching}

Each retained test session has one required action. The matcher groups tool calls by assistant turn, so parallel tool calls in a single assistant message are evaluated as one batch. It scans the trajectory until it finds the first batch containing the required tool; later batches are ignored once this batch decides the verdict.

For \texttt{tool\_identity} rules, any call to the required tool in that first matching batch is sufficient. For \texttt{param\_id}, \texttt{param\_enum}, and \texttt{confirm} rules, every call to the same required tool in the first matching batch must match the specified comparison arguments; this rejects same-turn shotgun behavior such as calling the same mutate tool with both correct and incorrect targets. Extra calls to different tools in the same batch are ignored. Scalar arguments use exact equality, list arguments use set equality, and \texttt{confirm} rules compare \texttt{target\_params} by safe typed fields of the target tool.

\section{Detailed Experimental Settings}
\label{sec:appendix_setup}

This section details the models evaluated as the agent under test, their inference configuration, and the sweep execution settings. Table~\ref{tab:app_model_inventory} lists backend identifiers and exact official source pages. We cite the most specific public source available for each backend: official technical reports where public, otherwise official system cards, model cards, release pages, or API documentation; we do not treat broader family reports as exact technical reports for proprietary API models.

\paragraph{Model access terms.} The evaluated frontier models are proprietary API services rather than redistributed model weights. We access them through their provider APIs under the corresponding provider terms, and Table~\ref{tab:app_model_inventory} lists the official source page used for each model artifact.

\paragraph{Sweep and execution.} The main sweep crosses the 8 evaluated agents, the 4 variants (Table~\ref{tab:variants}), and the 20-persona cohort, giving $8 \times 4 \times 20 = 640$ (model, variant, persona) cells. Each cell runs one learning sequence followed by all rule-bound test sessions for that persona. Sessions are single-trial with a 20-turn cap, which serves as the max-step limit referenced in Appendix~\ref{app:evaluation}; primary metrics are computed per persona and macro-averaged as defined there. The sweep runs at a cell-level concurrency of ten with intra-cell test-session concurrency of one; scaffold backends and shared infrastructure are as in \S\ref{sec:exp_setup}.

\paragraph{Per-model reasoning configuration.} For the agent under test, we enable each provider's reasoning or thinking mode where exposed, with the per-model settings below. GPT-5.4 uses \texttt{reasoning\_effort=high}; Claude Opus 4.7 uses \texttt{output\_config.effort=high} with adaptive thinking enabled; Gemini 3 Flash Preview and Gemini 3.1 Pro Preview use \texttt{thinkingLevel=high}; DeepSeek V4 Pro and Flash use \texttt{thinking=\{type:enabled, reasoning\_effort:high\}}; Qwen3.6-Plus uses the boolean toggle \texttt{enable\_thinking=true}; MiniMax M2.7 exposes no reasoning toggle and always interleaves thinking. The user simulator, Router, and Classifier use each provider's default and do not enable reasoning.

\begin{table}[t!]
\centering
\footnotesize
\setlength{\tabcolsep}{2.6pt}
\renewcommand{\arraystretch}{1.04}
\begin{tabularx}{\columnwidth}{@{}l l X@{}}
\toprule
\textbf{Display name} & \textbf{Backend id} & \textbf{Src.} \\
\midrule
GPT-5.4 & \texttt{gpt-5.4} & \href{https://developers.openai.com/api/docs/models/gpt-5.4}{API docs} \\
Claude Opus 4.7 & \texttt{claude-opus-4-7} & \href{https://www.anthropic.com/news/claude-opus-4-7}{rel.}; \href{https://www.anthropic.com/system-cards}{card} \\
Gemini 3 Flash Preview & \texttt{gemini-3-flash-preview} & \href{https://ai.google.dev/gemini-api/docs/models/gemini-3-flash-preview}{API}; \href{https://storage.googleapis.com/deepmind-media/Model-Cards/Gemini-3-Flash-Model-Card.pdf}{card} \\
Gemini 3.1 Pro Preview & \texttt{gemini-3.1-pro-preview} & \href{https://ai.google.dev/gemini-api/docs/models/gemini-3.1-pro-preview}{API}; \href{https://deepmind.google/models/model-cards/gemini-3-1-pro/}{card} \\
Qwen3.6-Plus & \texttt{qwen3.6-plus} & \href{https://www.qwencloud.com/models/qwen3.6-plus}{API docs} \\
MiniMax M2.7 & \texttt{MiniMax-M2.7} & \href{https://www.minimax.io/news/minimax-m27-en}{rel.}; \href{https://huggingface.co/MiniMaxAI/MiniMax-M2.7}{card} \\
DeepSeek V4 Pro & \texttt{deepseek-v4-pro} & \href{https://huggingface.co/deepseek-ai/DeepSeek-V4-Pro}{report} \\
DeepSeek V4 Flash & \texttt{deepseek-v4-flash} & \href{https://huggingface.co/deepseek-ai/DeepSeek-V4-Flash}{report} \\
\bottomrule
\end{tabularx}
\caption{Evaluated API model inventory. The main paper uses display names; exact backend identifiers and official provider pages are reported here for reproducibility.}
\label{tab:app_model_inventory}
\end{table}

\section{Compute Budget and Inference Cost}
\label{app:compute_budget}

We report compute budget across the two LLM-using phases of the project, benchmark construction and the main 640-cell evaluation sweep, under two billing scenarios: a no-cache list price that bills every input token at the provider's standard input rate, and a with-cache cost that bills cached input tokens at the provider's cache-hit rate where supported. Aggregate phase-level numbers are in Table~\ref{tab:app_cost_summary}. Dollar figures use the per-million-token rates posted on the inference platform at the time of the sweep; cache-hit rates are read from per-cell token accounting written by the runner.

\paragraph{Benchmark construction.} Producing the 20-persona dataset, including persona ingest, standing-rule generation and QC, test-session generation with iterative bound checks, and learning-session skeleton and fill, consumed about 29\,M input and 15\,M output tokens across about 5{,}200 LLM calls. Authoring calls use GPT-5.4 and dominate dollar cost; QC trace simulation uses Gemini 3 Flash Preview. Test-session generation alone contributes about two-thirds of the construction token budget because each candidate rule passes through three rounds of upper-bound and lower-bound checks. Total construction cost is about \$140, or about \$7 per persona; this phase issues one-shot calls without prompt-cache reuse.

\paragraph{Main sweep.} The 640-cell sweep (8 models, 20 personas, 4 variants) executed 13{,}632 learning sessions and 9{,}088 test sessions against the agent under test, plus the corresponding calls to the user simulator, Router, and Classifier. Aggregate agent-under-test input volume is on the order of 2.2\,B tokens with about 23\,M output, the low output-to-input ratio reflecting that most cells are tool-call dominated. The user simulator and the scaffold are pinned to fixed backends (GPT-5.4 for the user simulator and the Classifier; Gemini 3 Flash Preview for the Router) and add a roughly constant 300\,M input and 15\,M output tokens across all eight evaluated agents, so the scaffold cost does not bias the per-model comparison.

\begin{table}[t!]
\centering
\footnotesize
\setlength{\tabcolsep}{3.0pt}
\renewcommand{\arraystretch}{1.05}
\resizebox{\columnwidth}{!}{%
\begin{tabular}{@{}lrrcrr@{}}
\toprule
\textbf{Phase / role} & \textbf{In (M)} & \textbf{Out (M)} & \textbf{Hit} & \textbf{\$ list} & \textbf{\$ cached} \\
\midrule
Benchmark construction & 29    & 15.0 & --     & 138   & 138 \\
\midrule
Sweep, agent under test  & 2{,}208 & 23.0 & 64\,\% & 3{,}114 & 1{,}394 \\
Sweep, scaffold roles    & 299    & 15.0 & 20\,\% & 298   & 196 \\
\midrule
\textbf{Total}            & \textbf{2{,}536} & \textbf{53.0} & 60\,\% & \textbf{3{,}550} & \textbf{1{,}728} \\
\bottomrule
\end{tabular}
}
\caption{Aggregate compute budget across the benchmark-construction and main-sweep phases. In and Out are total input and output tokens in millions. Hit is the fraction of input tokens served from provider prompt cache, averaged across the constituent calls. \$~list bills every input token at the provider's standard input rate; \$~cached bills cached input tokens at the provider's cache-hit rate where supported. Construction uses one-shot calls and has no cache reuse.}
\label{tab:app_cost_summary}
\end{table}

\paragraph{Cache and total cost.} Six of the eight evaluated agents and two of the three scaffold roles serve a measurable fraction of their input tokens from provider prompt cache, with measured cache-hit rates ranging from 36\,\% (Gemini 3.1 Pro Preview) to 92\,\% (DeepSeek V4 Flash); the Classifier hits cache at about 83\,\% because its prompt is short and reused across rule-routing calls. Across the full sweep, about 1.4\,B of 2.2\,B agent input tokens hit cache, a sweep-level hit rate of roughly 64\,\%. Aggregate project cost is about \$3{,}550 at posted list price and about \$1{,}730 after applying cache-hit discounts, with Claude Opus 4.7 and GPT-5.4 contributing about 60\,\% of the with-cache total. Wall-clock for the sweep was about 36 hours under a cell-level concurrency of ten and an intra-cell test-session concurrency of one.

\clearpage
\section{Additional Results}
\label{sec:appendix_results}

\subsection{Run Completion and Health}
\label{app:run_health}

All 640 planned cells in the 20-persona $\times$ 8-model $\times$ 4-variant sweep completed, with no missing learning or test sessions. We separate this completion check from runtime health: Table~\ref{tab:app_termination_health} audits normal versus abnormal session termination, and the paragraph below summarizes learning-session channel repair and scaffold integrity.

\begin{center}
\begin{minipage}{\columnwidth}
\centering
\footnotesize
\setlength{\tabcolsep}{3.8pt}
\renewcommand{\arraystretch}{1.05}
\begin{tabular}{@{}lrrrr@{}}
\toprule
\textbf{Variant} & \textbf{LS norm.} & \textbf{LS abn.} & \textbf{TS norm.} & \textbf{TS abn.} \\
\midrule
Default & 4502/4544 & 42 & 2198/2272 & 74 \\
ATR & 4504/4544 & 40 & 2200/2272 & 72 \\
Always & 4510/4544 & 34 & 2203/2272 & 69 \\
Oracle & -- & -- & 2233/2272 & 39 \\
\bottomrule
\end{tabular}
\captionof{table}{Runtime termination health over completed cells. LS normal denotes user-simulator \texttt{task\_complete}; TS normal denotes agent \texttt{finish\_session}. Abnormal terminations include max-step exits, no-progress exits, and empty-response exits. Oracle has no learning sequence.}
\label{tab:app_termination_health}
\end{minipage}
\end{center}

The user-channel scaffold did not create missing cells or missing classifier decisions. Across learning variants, text-turn rescue recovered at least 99.0\% of initially off-protocol user-facing outputs, duplicate \texttt{send\_to\_user} events occurred once, and Classifier runtime errors were zero. These channel-health diagnostics support the integrity of the acquisition metrics but are not used as evaluation outcomes.

\subsection{Repeated-Trial Stability}
\label{app:stability}

To assess stability, we reran all four variants on the 20-persona panel three times for Gemini 3 Flash Preview and DeepSeek V4 Flash, for a total of 480 cells. Table~\ref{tab:app_stability} reports the results alongside the original sweep.

\begin{center}
\begin{minipage}{\columnwidth}
\centering
\scriptsize
\setlength{\tabcolsep}{2.2pt}
\renewcommand{\arraystretch}{1.04}
\resizebox{\columnwidth}{!}{%
\begin{tabular}{@{}lrrrrr@{}}
\toprule
\textbf{Variant} & \textbf{Main} & \textbf{Run 1} & \textbf{Run 2} & \textbf{Run 3} & \textbf{Mean $\pm$ SD} \\
\midrule
\multicolumn{6}{@{}l}{\emph{DeepSeek V4 Flash}} \\
\texttt{default}     & 23.7 & 26.5 & 24.9 & 24.8 & $25.4 \pm 0.96$ \\
\texttt{atr}         & 23.2 & 23.9 & 20.1 & 25.6 & $23.2 \pm 2.83$ \\
\texttt{always\_ask} & 26.5 & 32.9 & 31.3 & 30.6 & $31.6 \pm 1.19$ \\
\texttt{oracle}      & 93.0 & 89.2 & 87.8 & 88.6 & $88.5 \pm 0.66$ \\
\textbf{Gap}         & \textbf{69.3} & \textbf{62.7} & \textbf{62.9} & \textbf{63.8} & $\mathbf{63.1 \pm 0.61}$ \\
\midrule
\multicolumn{6}{@{}l}{\emph{Gemini 3 Flash Preview}} \\
\texttt{default}     & 15.6 & 13.5 & 15.9 & 18.2 & $15.9 \pm 2.35$ \\
\texttt{atr}         & 26.6 & 24.4 & 23.3 & 26.6 & $24.8 \pm 1.67$ \\
\texttt{always\_ask} & 25.8 & 24.9 & 27.2 & 24.6 & $25.6 \pm 1.43$ \\
\texttt{oracle}      & 91.0 & 91.1 & 91.3 & 91.5 & $91.3 \pm 0.19$ \\
\textbf{Gap}         & \textbf{75.4} & \textbf{77.6} & \textbf{75.4} & \textbf{73.3} & $\mathbf{75.4 \pm 2.16}$ \\
\bottomrule
\end{tabular}}
\captionof{table}{Repeated-trial TSAcc (\%, 20-persona macro). Main is the original sweep; Mean and sample SD are computed from unrounded values for Runs 1--3. Gap is \texttt{oracle}$-$\texttt{default}.}
\label{tab:app_stability}
\end{minipage}
\end{center}

Across the three reruns, the sample SD values range from 0.19 to 2.83 points, and Gap remains at least 62.7 points in all six model--run pairs. These reruns cover two of the eight evaluated models.

\subsection{Acquisition Diagnostics}
\label{app:acquisition_diagnostics}

Table~\ref{tab:app_acq_full} reports the full per-model acquisition diagnostics summarized in \S\ref{sec:exp_diagnostics}. Ask is RuleAsk, measuring how often the agent asks standing-rule questions during learning sessions; Cov is RuleCov, measuring how much of the persona's retained rule pool is acquired; App is AppliedRate, measuring whether learnable, classifier-confirmed acquired rules are later applied successfully; and GapRec is the fraction of the \texttt{default}-to-\texttt{oracle} gap recovered by the non-oracle variant.

\begin{center}
\begin{minipage}{\columnwidth}
\centering
\scriptsize
\setlength{\tabcolsep}{1.6pt}
\renewcommand{\arraystretch}{1.03}
\resizebox{\columnwidth}{!}{%
\begin{tabular}{@{}lrrrr@{}}
\toprule
\textbf{Model} & \textbf{Ask}$\downarrow$ & \textbf{Cov}$\uparrow$ & \textbf{App}$\uparrow$ & \textbf{GapRec}$\uparrow$ \\
\midrule
\multicolumn{5}{@{}l}{\emph{ATR}} \\
GPT-5.4           & 0.01 & 0.4 & 0.0 & -2.2 \\
Claude Opus 4.7   & 0.14 & 7.1 & 97.7 & 9.2 \\
Gemini 3 Flash Preview    & 0.51 & 12.0 & 78.4 & 14.6 \\
Gemini 3.1 Pro Preview    & 0.80 & 13.3 & 88.9 & 15.5 \\
Qwen3.6-Plus      & 0.01 & 0.7 & 100.0 & -1.2 \\
MiniMax M2.7      & 0.00 & 0.0 & -- & 1.3 \\
DeepSeek V4 Pro   & 0.01 & 0.6 & 50.0 & 1.8 \\
DeepSeek V4 Flash & 0.05 & 2.1 & 100.0 & -0.7 \\
\midrule
\textbf{AVG ATR} & \textbf{0.19} & \textbf{4.5} & \textbf{73.6} & \textbf{4.8} \\
\midrule
\multicolumn{5}{@{}l}{\emph{Always}} \\
GPT-5.4           & 1.00 & 12.7 & 80.9 & 13.5 \\
Claude Opus 4.7   & 1.01 & 15.6 & 83.2 & 14.1 \\
Gemini 3 Flash Preview    & 1.01 & 14.4 & 70.0 & 13.5 \\
Gemini 3.1 Pro Preview    & 1.00 & 11.3 & 87.2 & 8.4 \\
Qwen3.6-Plus      & 0.96 & 10.3 & 73.5 & 6.8 \\
MiniMax M2.7      & 0.09 & 1.8 & 25.0 & -3.6 \\
DeepSeek V4 Pro   & 0.98 & 14.5 & 70.6 & 5.6 \\
DeepSeek V4 Flash & 0.97 & 13.3 & 71.4 & 4.0 \\
\midrule
\textbf{AVG Always} & \textbf{0.88} & \textbf{11.7} & \textbf{70.2} & \textbf{7.8} \\
\bottomrule
\end{tabular}
}
\captionof{table}{Full per-model acquisition diagnostics (20-persona macro). Ask = RuleAsk, measured per learning session; Cov = RuleCov; App = AppliedRate over learnable, classifier-confirmed acquired rules. Cov, App, and GapRec are reported in \%.}
\label{tab:app_acq_full}
\end{minipage}
\end{center}

\subsection{Capability Breakdown}
\label{app:capability_breakdown}

Aggregate TSAcc hides where the gap is concentrated. We therefore break Table~\ref{tab:main} down by tool domain and by action-surface type.

\begin{figure*}[t]
\centering
\includegraphics[width=.94\textwidth]{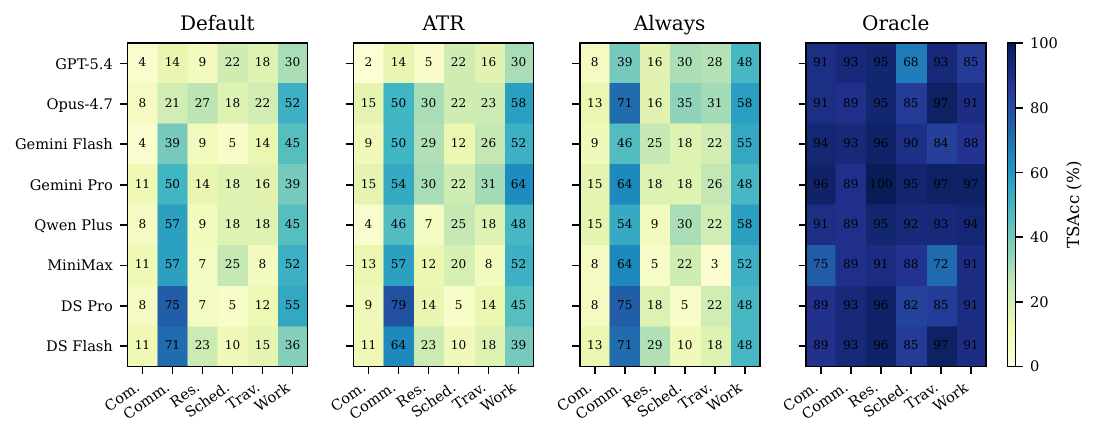}
\caption{Domain-level TSAcc shows broad executable upper bounds but uneven non-oracle behavior. Each heatmap cell is TSAcc (\%) aggregated over all 20 personas for one completed model, variant, and tool domain.}
\label{fig:per_domain}
\end{figure*}

\begin{figure*}[t]
\centering
\includegraphics[width=.94\textwidth]{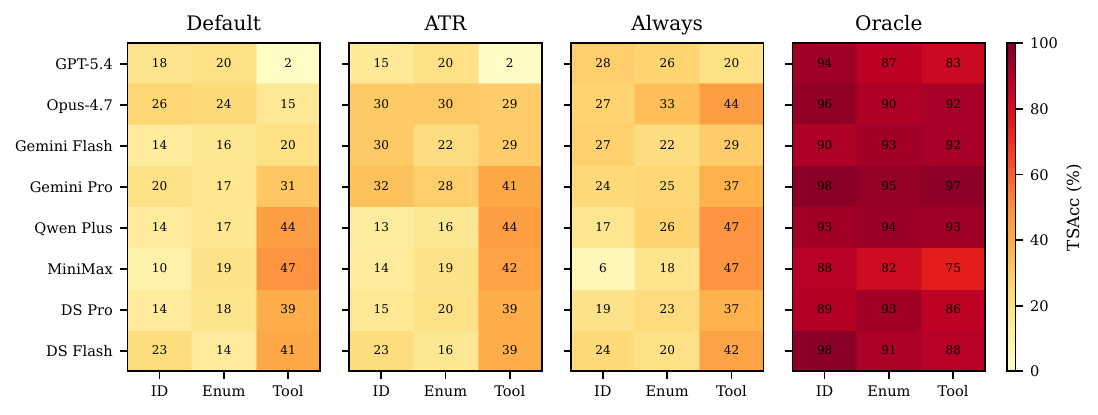}
\caption{Action-surface breakdown for the three non-sparse check types. ID = \texttt{param\_id}, Enum = \texttt{param\_enum}, and Tool = \texttt{tool\_identity}; confirmation rules are omitted from this figure because Table~\ref{tab:bench_stats} shows that bucket is too small for stable accuracy claims.}
\label{fig:check_type}
\end{figure*}

\textbf{Appendix Finding A1: the oracle gap is broad across domains and action surfaces.} Figure~\ref{fig:per_domain} shows that \texttt{oracle} reaches 67.5--100\,\% across completed (model, domain) cells, while the three learning-sequence variants remain much lower. Commerce is hardest without rule knowledge: across \texttt{default}, \texttt{atr}, and \texttt{always\_ask}, completed models stay between 1.9 and 15.1\,\% TSAcc. Communication and workspace are partially solvable even without acquisition, reaching 14.3--78.6\,\% and 30.3--63.6\,\%, respectively, which suggests that some domain decisions are recoverable from task-local cues or generic model priors. The benchmark therefore does not expose a single uniform failure; it separates domains where hidden standing rules dominate from domains where rule-naive behavior can sometimes succeed.

Figure~\ref{fig:check_type} shows the same pattern by action surface. \texttt{oracle} reaches 74.6--97.6\,\% on \texttt{param\_id}, \texttt{param\_enum}, and \texttt{tool\_identity}, confirming that all three action surfaces are executable when the rule answer is known. In contrast, non-oracle cells stay below 33\,\% for \texttt{param\_id} and at or below 33.3\,\% for \texttt{param\_enum}, while \texttt{tool\_identity} is higher but still capped at 47.5\,\%. Future ATR methods therefore need both better elicitation and better binding of acquired natural-language rules to concrete tool arguments.

\paragraph{Numeric summaries.} Table~\ref{tab:app_domain_summary} provides compact numeric summaries for the heatmaps in Figures~\ref{fig:per_domain} and~\ref{fig:check_type}. Each entry is TS-micro accuracy over completed cells, pooling all personas and evaluated models for the corresponding variant and bucket. Table~\ref{tab:app_persona_summary} reports the complementary persona-macro view, averaging across the eight evaluated models for each persona.

\begin{center}
\begin{minipage}{\columnwidth}
\centering
\footnotesize
\setlength{\tabcolsep}{3.0pt}
\renewcommand{\arraystretch}{1.05}
\emph{By domain}

\vspace{0.15\baselineskip}

\begin{tabular}{@{}lrrrrrr@{}}
\toprule
\textbf{Variant} & \textbf{Com.} & \textbf{Comm.} & \textbf{Res.} & \textbf{Sched.} & \textbf{Trav.} & \textbf{Work} \\
\midrule
Default & 8.0 & 48.2 & 13.2 & 15.0 & 15.2 & 44.3 \\
ATR & 9.9 & 51.8 & 19.0 & 17.5 & 19.1 & 48.5 \\
Always & 11.1 & 60.7 & 17.0 & 20.9 & 21.3 & 51.9 \\
Oracle & 89.4 & 91.1 & 95.5 & 85.6 & 89.9 & 90.9 \\
\bottomrule
\end{tabular}

\vspace{0.55\baselineskip}

\emph{By action surface}

\vspace{0.15\baselineskip}

\setlength{\tabcolsep}{5.5pt}
\begin{tabular}{@{}lrrrr@{}}
\toprule
\textbf{Variant} & \textbf{ID} & \textbf{Enum} & \textbf{Tool} & \textbf{Confirm} \\
\midrule
Default & 17.1 & 18.3 & 29.9 & 0.0 \\
ATR & 21.4 & 21.6 & 33.1 & 6.2 \\
Always & 21.6 & 24.4 & 38.1 & 12.5 \\
Oracle & 92.9 & 90.5 & 88.1 & 12.5 \\
\bottomrule
\end{tabular}
\captionof{table}{Capability TSAcc summaries (\%, TS-micro). The upper block groups by domain and the lower block groups by action-surface type. The confirm bucket has only two retained rules.}
\label{tab:app_domain_summary}
\end{minipage}
\end{center}

\begin{table*}[t]
\centering
\scriptsize
\setlength{\tabcolsep}{3.3pt}
\renewcommand{\arraystretch}{1.02}
\begin{tabular}{@{}lrrrrr@{\hspace{1.2em}}lrrrrr@{}}
\toprule
\textbf{Persona} & \textbf{Def.} & \textbf{ATR} & \textbf{Always} & \textbf{Oracle} & \textbf{Gap} &
\textbf{Persona} & \textbf{Def.} & \textbf{ATR} & \textbf{Always} & \textbf{Oracle} & \textbf{Gap} \\
\midrule
anna\_strahan & 17.3 & 20.2 & 22.1 & 97.1 & 79.8 &
joseph\_eldridge & 33.1 & 31.6 & 39.7 & 89.7 & 56.6 \\
cassandra\_tovar & 9.1 & 9.1 & 13.6 & 84.1 & 75.0 &
maximo\_esteban\_irizarry & 12.5 & 15.9 & 17.0 & 92.0 & 79.5 \\
cecilia\_lisette\_garciaperez & 15.6 & 17.7 & 19.8 & 92.7 & 77.1 &
michael\_argo & 28.7 & 31.6 & 33.1 & 91.2 & 62.5 \\
charlie\_james & 14.6 & 19.4 & 20.1 & 81.2 & 66.7 &
nancy\_marr & 15.2 & 24.1 & 21.4 & 86.6 & 71.4 \\
eduardo\_rivera & 20.6 & 25.0 & 22.1 & 93.4 & 72.8 &
rafael\_ortiz & 17.0 & 22.7 & 22.7 & 98.9 & 81.8 \\
esther\_hoitsma & 9.8 & 16.1 & 17.0 & 85.7 & 75.9 &
romisha\_shields & 25.9 & 30.4 & 37.5 & 84.8 & 58.9 \\
gabriel\_campbell & 24.2 & 26.7 & 29.2 & 90.0 & 65.8 &
roosevelt\_brown & 15.8 & 20.8 & 27.5 & 98.3 & 82.5 \\
iris\_parker & 18.8 & 20.8 & 20.8 & 92.7 & 74.0 &
samantha\_bello & 10.8 & 16.7 & 18.3 & 87.5 & 76.7 \\
jennifer\_philmon & 34.4 & 29.2 & 37.5 & 97.9 & 63.5 &
theodora\_tong & 27.6 & 36.2 & 38.2 & 88.2 & 60.5 \\
john\_brown & 22.9 & 30.2 & 29.2 & 94.8 & 71.9 &
vanessa\_davis & 19.2 & 20.0 & 20.8 & 90.0 & 70.8 \\
\bottomrule
\end{tabular}
\caption{Per-persona TSAcc summary (\%, persona-macro across the eight evaluated models). Gap is \texttt{oracle} minus \texttt{default}.}
\label{tab:app_persona_summary}
\end{table*}

\FloatBarrier

\clearpage
\section{Worked Trajectory Examples}
\label{sec:appendix_cases}

This section gives concrete, trace-level examples rather than an aggregate
failure analysis. Each example is drawn from a full human-readable evaluation
trace; the left panel compresses the interaction, and the right panel records
the Router/Classifier or evaluator state.

\subsection{Learning-Session Examples}
\label{app:ls_examples}

\begin{center}
\begin{minipage}{\columnwidth}
\begin{tcolorbox}[
  title={LS Example 1: Question Hits a Retained Rule},
  boxsep=1pt,
  left=2pt,
  right=2pt,
  top=2pt,
  bottom=2pt
]
\footnotesize
\setlength{\smallskipamount}{2pt plus 1pt minus 1pt}
\begin{tabularx}{\linewidth}{@{}>{\hsize=1.12\hsize\raggedright\arraybackslash}X@{\hspace{0.8em}}>{\hsize=0.88\hsize\raggedright\arraybackslash}X@{}}
\textbf{Compressed interaction trace} & \textbf{Scaffold reading} \\
\midrule
\textbf{Source.} One \texttt{atr}/Gemini 3 Flash Preview learning session.

\smallskip
The user asks for help replying to a message thread. The agent calls
\texttt{search\_messages}, finds one thread, and asks the user what the reply
should say. The user provides reply points.

\smallskip
The agent calls \texttt{draft\_message}, asks for approval, receives approval,
and then calls \texttt{send\_draft}. After the task is complete, it asks
whether messages in this category should always be drafted first for review.
&
\textbf{Routed question.} The agent asks whether future messages of the same
category should be drafted for review before sending.

\smallskip
\textbf{Router.} \textsc{rule}, strict=true.

\smallskip
\textbf{Classifier.} \textsc{hit}:
\texttt{communication\_01}.

\smallskip
\textbf{Canonical user answer.} The user always wants editorial or culturally
sensitive messages drafted first so they can revise them before they go out.

\smallskip
\textbf{Close.} The simulator returns that answer, the agent acknowledges the
standing rule, and the LS ends.
\end{tabularx}
\end{tcolorbox}
\captionof{table}{A learning session where the post-task standing-rule question maps to
a retained rule.}
\label{tab:app_ls_hit_example}
\end{minipage}
\end{center}

\subsection{Test-Session Examples}
\label{app:ts_examples}

\begin{center}
\begin{minipage}{\columnwidth}
\begin{tcolorbox}[title={TS Example: Wrong Argument on the First Relevant Action}]
\footnotesize
\begin{tabularx}{\linewidth}{@{}>{\raggedright\arraybackslash}X@{\hspace{1.0em}}>{\raggedright\arraybackslash}X@{}}
\textbf{Compressed interaction trace} & \textbf{Evaluator reading} \\
\midrule
\textbf{Source.} One \texttt{atr}/Gemini 3 Flash Preview test session.

\smallskip
The user asks the agent to handle an invitation after checking the relevant
calendar for conflicts.

\smallskip
The agent calls \texttt{list\_events}, finds an overlapping event, later checks
unrelated windows, then makes the invitation-response call with
\texttt{response=accept} and later issues a rescheduling call.
&
\textbf{Required action.} The invitation-response tool.

\smallskip
\textbf{Observed first relevant action.} The required tool is present, but the
argument is wrong.

\smallskip
\textbf{Gold argument.} \texttt{response=tentative}.

\smallskip
\textbf{Actual argument.} \texttt{response=accept}.

\smallskip
\textbf{Verdict.} Gold fail. This illustrates the TS matching rule used
throughout ATRBench: a later repair does not matter if the first action touching
the rule's action surface carries the wrong decision.
\end{tabularx}
\end{tcolorbox}
\captionof{table}{A test session that fails even though the required tool appears,
because the first relevant call carries the wrong enum argument.}
\label{tab:app_ts_fail_example}
\end{minipage}
\end{center}

\clearpage
\section{Prompts}
\label{sec:appendix_prompts}

We reproduce the evaluation prompts below with placeholders for dynamic context.

\lstset{
  breaklines=true,
  xleftmargin=3pt,
  breakindent=0pt,
  columns=fullflexible,
  basicstyle=\ttfamily\scriptsize,
  aboveskip=4pt,
  belowskip=4pt
}

\tcbset{
  breakable,
  boxrule=1pt,
  colback=gray!5,
  colframe=gray!55!black,
  coltitle=black,
  fonttitle=\small\bfseries,
  left=4pt,
  right=4pt,
  top=4pt,
  bottom=4pt,
  before skip=8pt,
  after skip=10pt
}

{\small

\subsection{Data Construction Prompts}
\label{app:data_construction_prompts}

The seven prompts below drive the construction pipeline (Appendix~\ref{sec:appendix_construction}): standing-rule generation and two-axis QC, rule-bound test-session generation, QC trace simulation, and repair, and learning-session skeleton and fill. Double-brace placeholders (e.g.\ \texttt{\{\{RULE\_JSON\}\}}) are filled at call time; few-shot pools and persona or tool context are elided where marked. Non-ASCII glyphs in the deployed prompts are transcribed to ASCII here for typesetting.

\begin{tcolorbox}[title={Rule Generation (\texttt{rules/gen})}]
\begin{lstlisting}
# Rule Generation

From the persona below, produce {{N_RULES}} standing rules. Each rule is one sentence describing a long-term preference of this user and must map cleanly to one tool call from the tool list.

## What makes a good rule

1. **Grounded in the persona.** Point to a specific phrase, behavior, activity, occupation, or community. Generic adjectives ("creative", "diligent", "careful") do not count -- they fit anyone.

2. **Surprising default.** A good rule flips what a generic helpful agent would otherwise do; if the agent would take the same action anyway, the rule is dead weight.

   Today's LLM agents have a strong cautious / non-destructive bias, so cautious-side wedges (archive-over-delete, pause-over-cancel, confirm-before-destructive-write, prioritize-important, track-after-ordering) are NOT surprising -- they are the default.

   Example: "Delete promo emails outright." (Agents default to archive; this rule pushes the destructive action.)

3. **Maps to one tool call.** See `check_type` below.

Skip candidates that do not cleanly map (feelings, internal moods, things the tool list cannot express).

## Output

JSON list. Each entry:

```json
{
  "rule_text": "<third-person sentence; do not name tools or parameters>",
  "canonical_answer": "<first-person sentence that stands on its own (e.g., 'I always X for Y' / 'When Y, I do Z' / 'For X, I prefer Y'); do not name tools or parameters>",
  "counterfactual_default": "<what someone without this rule would naturally do>",
  "evidence": [
    {"type": "direct",   "content": "<exact phrase from the persona>"}
    // OR {"type": "inferred", "content": "<reasoning chain rooted in specific persona facts>"}
  ],
  "check_type": "tool_identity | param_id | param_enum | confirm",
  "action_step": {
    "tool":  "<exact tool name from tool list>",
    "param": "<param name on that tool, OR null, OR -- for confirm rules -- the inner mutate tool name>"
  }
}
```

## `check_type` -- pick one of four

- **`tool_identity`** -- pins WHICH TOOL. `param = null`.
- **`param_enum`** -- pins an ENUM VALUE on an enum-typed param.
- **`param_id`** -- pins WHICH ITEM (by attribute) on a `*_id` / `*_ids` parameter. Not for ids the user hands in ("delete file X" -- that is instruction following).
- **`confirm`** -- asks for `get_user_confirmation` before a mutate; `param` carries the INNER mutate tool name (NOT a parameter on `get_user_confirmation`). A confirm rule may also pin an inner-param preference; that surfaces later as test-session `gold_value`, not in `action_step`.

## Hard rules

- Tool and param names are exact strings from the tool list.
- `param` is what the rule pins (output choice), not what the user hands in (input id). For instance, pin `selected_alternative_stop_ids`, not `disrupted_item_id`.
- Do not put a *rejected* tool in `action_step`. For "draft instead of send", `tool=draft_message`, never `send_message` -- the rejected tool belongs in `counterfactual_default`.
- For "do X, don't do Y" rules, X and Y must be alternatives -- calling X precludes Y, or Y is something nobody would call here.

  Example: "archive instead of delete" -- archive and delete are exclusive cleanup actions.

  Counter-example: "label community emails, don't archive" -- labeling and archiving are independent steps, so the agent does both anyway.

- No duplicate or contradicting bindings. "Direction" = which side the rule prefers. At most one rule per `(tool, param, direction)`. Two rules on the same `(tool, param)` are OK only if their directions are orthogonal axes (one pins cuisine, another pins seating -- both can hold on one restaurant call). Two rules pinning the same trigger to opposite values: drop the weaker one.

### Tool pairs that look similar -- pick deliberately

- `plan_trip` (new) vs `replan_trip` (existing trip; needs `trip_id` + `disrupted_item_id`).
- `modify_event` (anything except time) vs `reschedule_event` (time only).
- `update_tracker` (append to existing; needs `tracker_id`) vs `create_document` (new doc / spreadsheet / note).

## Examples

```json
// tool_identity
{"rule_text": "When cleaning up work files, archive rather than delete.",
 "check_type": "tool_identity",
 "action_step": {"tool": "archive_files", "param": null}}

// param_enum
{"rule_text": "When an invite conflicts with something unresolved, respond tentatively rather than declining.",
 "check_type": "param_enum",
 "action_step": {"tool": "respond_to_event_invite", "param": "response"}}

// param_id
{"rule_text": "For garden purchases, favor locally-adapted native plants over generic landscaping options.",
 "check_type": "param_id",
 "action_step": {"tool": "place_order", "param": "product_id"}}

// confirm
{"rule_text": "When new fad-style health services come up, check with her before booking.",
 "check_type": "confirm",
 "action_step": {"tool": "get_user_confirmation", "param": "book_service_appointment"}}
```

## Tool list

{{TOOLS_TEXT}}

## Persona

{{STRUCTURED_PERSONA}}

## Final per-rule checks

1. `counterfactual_default` names a *different action*, not just less of the same.
2. The same persona trait is not already covered by another rule in your output.

Output JSON list only. No code fence. No other text.
\end{lstlisting}
\end{tcolorbox}

\begin{tcolorbox}[title={Rule Quality Control (\texttt{rules/qc})}]
\begin{lstlisting}
# Rule QC

Score each rule on two independent axes (`counter_default` and `binding_sound`). The downstream keep / remove tag AND-s them -- you only emit the per-axis labels and reasons. Score each rule independently; you do not need to consolidate duplicates, contradictions, or evidence quality across rules (all handled elsewhere). Score using `rule_text`, `counterfactual_default`, `check_type`, and `action_step`.

**Terminology.** *passive* = a generic helpful agent that does NOT know this rule. Note that `rule_text` is pipeline metadata; passive does NOT see it -- at test time passive sees a neutral instruction generated from the rule + tool list + ontology defaults, so judge passive from the underlying scenario rather than from `rule_text` wording. *gold action* = the tool call this rule wants the agent to take (i.e. `action_step`).

---

## `counter_default` (yes | no)

> If we strip this rule away, would a generic helpful agent in the same scenario take the gold action anyway?

Yes -> `counter_default = no` (rule is redundant, no personalized wedge). No -> `counter_default = yes` (real wedge).

**NOT real wedges** (mark `no`). Current LLMs have a strong cautious / non-destructive default, so cautious-side rules are what passive does anyway: any "archive / pause / preserve / keep" rule that picks the safer option, and any "confirm before a destructive write" rule.

**OK wedges** (do NOT mark `no`). Rules that flip the passive default or pin a value passive would not reach for.

Example: "draft instead of send directly" -- passive defaults to sending the prepared message; the rule flips that.

Example: "respond=tentative on conflict" -- passive picks accept or decline; the rule pins the third enum value.

**Subtle failure.** Rule says "do X, don't do Y" but X and Y are not mutually exclusive -- passive does both, so gold X is called even without the rule -> `counter_default = no`.

Counter-example: "label community emails, don't archive" -- labeling and archiving are independent steps; agents do both anyway.

### Examples for this axis

{{COUNTER_DEFAULT_POOL}}

---

## `binding_sound` (yes | no)

> Knowing this rule, would an agent naturally land on this exact `(check_type, action_step)`?

Yes -> `binding_sound = yes`. No -> `binding_sound = no`.

**Looks right but fails** (mark `no`):

- Rule is a cross-tool choice ("reminder rather than a new calendar event") but the binding lands on a param INSIDE one tool (e.g. `set_reminder.cadence`) -- an agent following the rule would care about which tool, not which cadence.
- Rule talks about a dimension (time, formality, theme) but the binding is `tool_identity` with `param = null` -- no parameter slot carries the dimension, so calling the tool does not express the preference.

**Looks wrong but actually OK** (do NOT mark `no`):

- Cross-tool wedge bound to the gold side -- rule "prefer A over B", binding `A, tool_identity`. The rule's action IS calling A; the rejected B belongs in `counterfactual_default`, not in the binding.
- `param_id` rule whose dimension lives in candidate `attributes` -- rule talks about time / theme / type, binding is `*_id`. The runtime populates each candidate's `attributes` with that dimension and the agent selects by reading them, so the binding carries the dimension indirectly.

**Enum catch-all warning.** Avoid `param_enum` whose enum includes an `"all"`-style catch-all. Agent safety bias favors the catch-all over the rule's specific value, making the rule unreliably testable.

### Examples for this axis

{{BINDING_SOUND_POOL}}

---

## Rules to review

```json
{{RULES_JSON}}
```

## Output format

```json
[
  {
    "rule_text": "<copy verbatim from input>",
    "counter_default": "yes" | "no",
    "counter_default_reason": "<one line; if 'no', name what passive would do by default; if 'yes', write 'passes'>",
    "binding_sound": "yes" | "no",
    "binding_sound_reason": "<one line; if 'no', name what the rule asks for vs what the binding lands on; if 'yes', write 'passes'>"
  }
]
```

Output count = input count. Every input rule appears exactly once.

Output JSON list only. No code fence. No other text.
\end{lstlisting}
\end{tcolorbox}

\begin{tcolorbox}[title={Test-Session Generation (\texttt{test\_sessions/gen})}]
\begin{lstlisting}
# Test Session Generation

Generate one test session for the given rule. At runtime the session is user-offline -- the agent executes alone, with no one to clarify with -- so the instruction must be self-contained and must NOT reveal the rule direction.

The session is valid iff the binary discrimination holds:

- **oracle** (knows the rule) hits gold tool + gold args.
- **passive** (does not know the rule), reading the same instruction + refs and reasoning by commonsense, does NOT hit gold.

Use `rule.counterfactual_default` as your prediction of passive's behavior and point your decoy design that way so the oracle / passive gap stays sharp.

This is a single-call task -- no refine loop. Reason through the steps below internally, then emit the final JSON.

---

## Locate the discriminative dimension

From `rule_text` + `canonical_answer` + `counterfactual_default`: what does oracle do (gold)? what does passive do (counterfactual)? where is the difference -- `param_id` (which ref selected) / `param_enum` (which enum value) / `tool_identity` (which tool)?

## Refs design

### Per-decoy pull (param_id only)

Plan 1 gold + >=2 decoys. Every decoy needs at least one concrete *pull attribute* -- a property that, on its own, attracts a rule-blind agent. Menu: lowest price / highest rating / most popular / official or authoritative / most recently updated / closest distance / most full-featured.

- Decoy check: "Without the rule, which attribute pulls toward this decoy?" If you cannot name it, redesign -- a parallel choice with no pull leaks because passive picks at random.
- Gold check: "With the rule, does the rule-aware agent uniquely lock on gold?" If a decoy also satisfies the rule + instruction demands, oracle could legitimately pick it -- redesign.

For `param_enum` / `tool_identity`, refs still need >=1 gold + 1-2 decoys for context, but gold determination depends on the enum value / tool, not on the ref set.

### Refs-as-a-whole must lean toward the counterfactual

Read `counterfactual_default`, work backwards to what signal in refs points that way, then amplify it. The rule-direction signal should be the *least* prominent.

- **tool_identity** (rule = `create_document`, counterfactual = `update_tracker`): refs must include 1-2 appendable-looking trackers so passive leans toward update.
- **param_enum** (rule = `scheme=by_project`): each ref carries the rule dimension AND other dimensions (`created_date`, `doc_type`) so passive has multiple classification paths.
- **param_id**: the rule-direction attribute must NOT be gold-exclusive. Give >=2 candidates with the rule's direction (1 gold + >=1 decoy), and let decoys beat gold on a secondary dimension (rating, distance, popularity) so passive picking by secondary signal lands on a decoy.

  Counter-example. Rule "same theme (nature)" + refs = 1 nature garden + 3 art / spa / cafe. Passive sorting "same theme" has no choice but gold.

  Example. Gold = garden (nature); decoy1 = overlook (nature, far); decoy2 = temple (cultural, 4.8 rating); decoy3 = cafe (4.9). Two nature candidates with secondary-dimension reversal.

Self-check: count refs carrying each rule-direction attribute. Only 1 (gold-exclusive) -> add a same-direction decoy.

### Two more ref traps

- **Hard-constraint exclusivity** (`param_id`). If instruction lists hard constraints ("Old Harbour + 2 adults + private bathroom"), every decoy must satisfy all of them, else instruction-filtering rejects them and gold becomes the only valid pick. Differentiate decoys on secondary dimensions, not on the hard constraints.
- **Project-lifecycle naming** (`param_enum scheme=by_project`). Ref names that read like one project's stages (survey -> costs -> recap -> packet) commonsense-pin passive on `by_project` even with neutral instruction. Mix names across multiple visible projects, vary `file_type`, and spread `created_date` over months.

## Gold triple-filter

All three must be yes before fixing gold:

1. **Rule-conformant**: gold satisfies every dimension of the rule direction. (Do not misread "A over B" -- gold must BE A, not just "not B".)
2. **Instruction-conformant**: gold satisfies the instruction's explicit non-rule demands (time, location, count, budget).
3. **Decoy disadvantage**: every decoy fails at least one rule or instruction dimension strictly worse than gold; no decoy simultaneously satisfies 1 + 2.

## Instruction

Write a natural one-paragraph user request. Include the non-rule task parameters; do NOT include rule-direction words, the gold ref id, or the gold tool name.

### Wording neutrality (the main passive-leak source)

For every concrete noun and verb, ask: "On its own, does this word push passive toward gold?" If yes, swap for neutral phrasing. Common leak patterns and the neutral rewrites:

- **Type-word** ("project files" when rule is `by_project`) -- neutralize to "random files that have piled up".
- **Action-word** ("draft a message" when rule is draft-over-send) -- neutralize to "send X a quick note".
- **Container / location-word** ("don't leave them in the workspace" when rule is archive-over-delete) -- neutralize to "clean them up".
- **Frequency / cadence-word** ("every week I do this" when rule is weekly cadence) -- neutralize to one absolute time ("Wednesday at 8pm"), and let oracle infer weekly from the rule.
- **Business-type word** ("claims, renewals, policy-change files" when rule is `by_project`) -- neutralize to "files I've collected over the past few months".
- **Replacement-need word** ("the compact one won't fit" when rule is exchange-over-refund and refs hold obvious replacements) -- neutralize to "I want to send this back".
- **Loaded-scenario** (long-haul "Seattle -> London, pick a seat" pre-commits passive to `window`; broken-item scenarios pre-commit `resolution_type='exchange'`; deadline-loaded messages pre-commit `priority='high'`; one-off scheduled tasks pre-commit `cadence='before_event'`) -- replace with a neutral scenario, or omit the cue field entirely.

Rule of thumb: as passive, read instruction alone -- does any word make you think "ah, must be option X"? If X = gold, drop the word. Legitimate task parameters (time, location, headcount, target person) must still be written clearly -- these are task context, not direction words. Test: remove the word; does the task still make sense? Yes -> keep; No -> it was a direction word, drop.

### Discoverability

For every ref id in `gold_value` (or in `*_id` / `*_ids` slots inside `gold_value` for confirm rules), the instruction must contain at least one token that loose-matches a token in one of the ref's searchable attributes (`name`, `title`, `subject`, `sender`, `location`, `region`, `city`, `address`, `destination`, `origin`, `cuisine`, `service_type`, `category`, `doc_type`, `tags`). Without this, the agent's search returns 0 results and the session is unsolvable.

Loose-match = after lowercasing and splitting on whitespace / punctuation, some instruction token equals or is a substring of some attribute token. If persona / refs use English place names while the instruction is Chinese (or vice versa), plant the English token verbatim somewhere in the instruction.

### Time handling

ATRBench does not model a time axis: search tools do NOT take time / date filters. The agent fetches refs first, then reads `date_time` / `start_time` / `departure_date` on each candidate.

- Default: no specific time in instruction ("book dinner for me").
- Relative time is forbidden: never write "next Thursday / this week" or any non-Latin-script equivalent -- runtime hides the current date.
- Absolute date is allowed only for time-related rules (cadence enum, time-triggered notification, scheduling-conflict pattern): use `YYYY-MM-DD`, and the gold ref's date field must contain this date as a substring (`"2026-05-18T15:00:00"` matches `2026-05-18`).

When in doubt, omit time.

### Derived entities

When the gold tool operates on a derived entity (orders / trips / reservations / bookings / appointments), the env hydrates the entity from the corresponding primary ref's attributes (see `derived:` sections in REF_SCHEMA for which primary ref carries which ID attribute). Place the derived ID on the right primary ref's `attributes`, else gold fails with "not found".

In the instruction, describe the entity by role / context, never by its specific name, and never with a literal id (`ord_xxx` / `rst_xxx`).

Example. "Return the candle I ordered last month." (Role.)

Counter-example. "Return the Cedar Glow Candle I ordered." (Specific name -- passive can match it directly.)

---

## Few-shot examples (same / similar bucket)

{{FEW_SHOT_EXAMPLES}}

---

## Counter-example -- real failure, do not repeat

`param_id` / numeric pull too obvious. Rule "prefer high-rating reliability over distance". Wrong refs: gold = Cascade Code (rating = 4.9, dist = 6.4 km); decoys top out at 4.7 rating with distances under 1 km. Passive locks onto "highest rating" = gold because LLMs are far more sensitive to rating jumps (4.7 -> 4.9) than to distance gaps (0.5 -> 6.4 km). Fix: mix gold's rating into the middle pack, and let a decoy own "highest rating + closest" so passive is pulled away.

---

## Output JSON schema

You write 3 fields (plus optional `list_match_mode`):

```json
{
  "instruction": "<one English paragraph; non-rule task params present, no rule-direction words>",
  "gold_value": <see dispatch table below>,
  "list_match_mode": "strict" | "subset",   // OPTIONAL -- only when gold_value is a list
  "references": [{"id": "...", "type": "...", "attributes": {...}}]
}
```

You do NOT write `domain`, `local_env.tools`, `session_id` / `session_type` / `persona_id` / `rule_id` / `rule_ref`, or `labels` -- the caller derives them.

### `gold_value` dispatch (by `rule.check_type` x param type)

Find `<rule.param>`'s type in the tool signature inside `DOMAIN_TOOLS`.

| `check_type` | `<rule.param>` type | `gold_value` |
|---|---|---|
| `tool_identity` | (n/a, no param) | `null` |
| `param_id` | `string` (single id) | `"<ref_id>"` |
| `param_id` | `array[string]` (list of ids) | `["<ref_id>", ...]` (typically length 1) |
| `param_enum` | `enum[...]` | `"<enum_value>"` |
| `confirm` | (param = inner mutate tool name) | `null` (confirm-only) OR `{<inner_param>: <inner_value>}` |

`list_match_mode` (emit only for `array[string]` `param_id` rules):

- `"strict"` (default): `set(gold) == set(actual)`, no extras. Use when rule intent is "only these qualify".
- `"subset"`: `set(gold) subseteq set(actual)`, gold ids must all appear and extras are tolerated. Use when rule intent is "always include these".

When in doubt, default to `"strict"` (you may omit the field).

### Confirm rule details

`rule.action_step.tool` is always `get_user_confirmation`; `rule.action_step.param` carries the inner mutate tool name (e.g. `"delete_files"`, `"book_service_appointment"`).

- No inner-param preference (purely "confirm vs not"): `gold_value = null`. Caller fills `arguments = {target_tool: <inner>}` and `compare_args = ["target_tool"]`.
- Inner-param preference (e.g. "ask before canceling the recently-added appointment"): `gold_value = {<inner_param>: <value>}`. Caller adds `target_params: <gold_value>` and `compare_args = ["target_tool", "target_params"]`. The inner param must exist on the inner mutate tool's signature with a safe type (`*_id` / `*_ids` / `enum` / `boolean`).

---

## Hard self-check (run before emitting)

1. Non-rule task params (location, headcount, target person) are written clearly in the instruction.
2. References contain exactly 1 object satisfying the rule direction for `param_id` (and inner-`param_id` for confirm rules).
3. `gold_value` matches `rule.check_type` x `<rule.param>` type per the dispatch table.
4. Ref attributes hold only neutral facts (numbers, categories, identity, time, participants) -- no value-judgments, recommendations, or actionability hints.

### Confirm rule special constraint

If `rule.check_type == "confirm"`, the instruction is an ordinary user request ("help me schedule X" / "handle it") and must NOT contain any wording about a confirm flow ("ask me first", "check with me", "confirm before", "get my permission", "decide later") or any two-step process narration ("first X, then Y"). The confirm step is what oracle infers from the rule -- if instruction leaks it, passive also calls `get_user_confirmation` and the discrimination disappears.

### id-reference hard red lines (static check enforces)

1. Any `*_id` / `*_ids` value in `gold_value` must be the id of an object in `references`. No fabricated ids.
2. `references[i].type` must be a primary type listed in REF_SCHEMA (the `type:` headers -- not the `derived:` sections).

---

## Inputs

### Rule

```json
{{RULE_JSON}}
```

### Domain = `{{DOMAIN}}` available tools

```json
{{DOMAIN_TOOLS}}
```

### Domain `{{DOMAIN}}` reference schema

{{REF_SCHEMA}}

---

Emit strict JSON now, no markdown code fence.
\end{lstlisting}
\end{tcolorbox}

\begin{tcolorbox}[title={Test-Session QC: Agent Trace Simulation (\texttt{test\_sessions/qc})}]
\begin{lstlisting}
# Test Session QC -- Agent Trace Simulation

You are an assistant. Complete the user's task by calling the available tools. Output the full trace in a single response -- list every tool call you intend to make, in order. Do not split across turns; do not wait for tool returns.

{{RULE_CONTEXT_BLOCK}}

## Task

### User request

{{INSTRUCTION}}

### Available tools

```json
{{TOOL_SPECS}}
```

### Objects in the environment

```
{{REFERENCES}}
```

> Candidate objects you may reference by id. Pick directly -- do NOT add a `search_*` / `list_*` discovery prologue.

## Conventions

- List tool calls in natural execution order.
- Argument values must conform to the schema: `enum` values come from the listed set (never invent); `boolean` is `true` / `false`; other `string` params must not be fabricated; `id` params may reference only the ids listed above.
- Always supply required parameters; for optional parameters, supply a value only when the user request, the environment objects, or an applicable user preference clearly determines it.

## Output format (strict JSON, no code fence)

```jsonc
{
  "trace": [{"tool": "<tool_name>", "arguments": {<kwargs>}}, ...],
  "reason": "one sentence explanation"
}
```

Output now.
\end{lstlisting}
\end{tcolorbox}

\begin{tcolorbox}[title={Test-Session Repair (\texttt{test\_sessions/refine})}]
\begin{lstlisting}
# Test Session Repair

The previous test session failed QC. Rewrite a new version (strict JSON, no code fence) such that, simultaneously:

- **Binary discrimination holds**: oracle hits gold; passive (commonsense) does NOT.
- The agent's first tool call hits the rule's relevant tool (the rule's decision point is not at a side step).
- Instruction is self-contained: every non-rule task parameter is written clearly.
- Refs hold only neutral facts (numbers / categories / source / identity / time / participants) -- no value-judgments, recommendations, or actionability hints.
- For `confirm` rules, the instruction is an ordinary user request and must not leak the confirm flow (no "ask me first", "check with me", "confirm before", or two-step process narration).

Repair is NOT patching -- you may change any field.

Cross-reference `failure_reasons` with the N `upper_reasons` / `lower_reasons` and their per-sample traces and match flags. Keywords repeated in passive self-reports point to where instruction or refs leak direction.

## Static-error precise fixes

- `instruction_leaks_ref_id:<ID>` -> replace the literal id string with a natural-language description ("the train ticket from Florence to Venice" instead of `gt_florence_venice_20260512`).
- `action_{i}_id_missing_from_refs:<tool>.<arg>=<id>` -> `gold_value` references an id not in `references`. Either add the object (`{id, type, attributes}`) or change `gold_value` to an existing id.
- `ref_type_invalid:<type>:domain=<domain>` -> swap to a primary type from REF_SCHEMA's `#### type:` headers (NOT the `derived:` sections, which are auto-hydrated).
- `action_step_tool_is_exploration_prefix` -> rule-level error; the repair stage cannot fix it -- return to rule generation.

## Reading the QC failure mode -> targeted fix

- **Mode A** (`upper_matched = False`): oracle did not hit gold. Causes: refs lack an object satisfying the rule direction (fix `references` so exactly 1 ref satisfies every dimension); `gold_value` is not what oracle would pick (fix `gold_value`); or the instruction's task parameters conflict with the gold ref's attributes ("Beijing restaurant" but gold ref is Shanghai -- align one with the other). Read `upper_reasons` for the sentence where oracle says why it picked X instead of gold.

- **Mode B** (`upper_matched = True, lower_matched = True`): the main failure mode -- passive also hit gold. Find repeated keywords in `lower_reasons`:

  - "I picked X because the instruction said Y" -> instruction leak. Apply the wording-neutrality guidance from the generation prompt (drop the type / action / container / cadence / business-type / replacement-need / loaded-scenario leak).
  - "I picked X because its attribute Z fits best" -> refs let gold's rule-direction attribute be too prominent. Apply the refs-as-a-whole guidance from the generation prompt: give decoys pull attributes (price / rating / popularity / official / newest / closest / most full-featured); for `param_id`, if gold is exclusive on the rule attribute, the decoys must beat gold on a secondary dimension; for `tool_identity`, refs must include candidates the counterfactual tool would handle.

- **Mode C** (`upper_matched = False, lower_matched = True`): both wrong. Usually gold or markers were set wrong -- fix oracle per Mode A; if the lower problem does not disappear as a side effect, apply Mode B.

## Real-failure reminders -- do not repeat

- `param_id` / gold-exclusive on the rule direction -> add 1-2 same-direction decoys that beat gold on a secondary dimension.
- `param_id` / numeric gap too obvious (gold rating 4.9 vs. decoys topping out at 4.7): LLMs lock onto "highest rating" = gold; gold rating must NOT be the maximum.
- `param_enum` / task-word leak ("project" ~= `by_project`; "key" ~= critical): neutralize phrasing.
- Rule direction == LLM default: fundamentally unconstructible -- do not force-fix, report "unconstructible".

## Few-shot examples (same / similar bucket)

{{FEW_SHOT_EXAMPLES}}

---

## Inputs

### Previous test session

```json
{{CURRENT_TASK}}
```

### Failure reasons

```json
{{FAILURE_REASONS}}
```

### Rule

```json
{{RULE_JSON}}
```

### Domain = `{{DOMAIN}}` available tools

```json
{{DOMAIN_TOOL_SPECS}}
```

### References format contract

{{REF_SCHEMA}}

---

## Hard MUSTs (violation -> entire repair is discarded)

Retain these 3 fields; values may change but keys cannot be dropped:

1. `instruction` (str, non-empty).
2. `references` (>=3 objects for `param_id`; smaller is OK for other check_types if there is at least 1 gold-context object).
3. `gold_value` (per `rule.check_type`):
   - `tool_identity` (mutate, `action_step.param = null`) -> `null`
   - `param_id` -> `"<gold ref id>"` (must exist in `references`)
   - `param_enum` -> `"<enum value>"` from the enum's value set
   - `confirm` -> `null` (confirm-only) or `{"<inner_param>": <value>}` (inner_param must be safe-typed on `rule.action_step.param`'s tool signature)

Do NOT output `domain`, `local_env.tools`, `labels`, or `session_id` -- the caller derives them. A common past mistake was emptying or mistyping `gold_value`; do not repeat it.

Output the complete repaired test session JSON now, no markdown code fence.
\end{lstlisting}
\end{tcolorbox}

\begin{tcolorbox}[title={Learning-Session Skeleton (\texttt{learning\_sessions/skeleton})}]
\begin{lstlisting}
# Learning-Session Skeleton Generation

You are a scenario architect for AI-agent evaluation data. Given a persona's structured profile, produce {{N}} learning-session skeletons. Each skeleton has two fields only:

- `domain` -- one of the 6 supported domains
- `theme_one_line` -- one English sentence describing what this persona, in everyday life, asks an AI agent to do

A downstream fill stage fills in the full session content (instruction, task_params, references, expected_tools). Your job is pool-level scenario architecture only -- you commit to *what* each session is about, not *how* the agent will execute it.

---

## Generation principles

1. **A daily-interaction trail.** The {{N}} entries together form a stretch of this persona's interactions with the agent. Order them in natural temporal order (earliest first); the caller stamps `day_offset` from the index. Adjacent entries may carry causal or topic continuity (booked weekend gathering -> drafted thank-you note; planned business trip -> booked flight -> booked hotel) -- 2-4 such chains is enough. Do NOT batch entries by domain (first 5 reservation, last 5 workspace) -- real life interleaves.

2. **Domain distribution follows the persona's actual life.** You are not required to cover all 6 domains. Read the persona's occupation, lifestyle, and social_context: an art enthusiast may have lots of commerce; a commuter office worker may lean on communication / scheduling; a homemaker may use reservation more. Irrelevant domains can be entirely absent -- do not shoehorn.

3. **Bake situational variety into theme text.** Since you emit one sentence per session, vary time-of-day, day-of-week, location, and mood through wording. Mix early-morning errands, weekend evenings, rushed weekday afternoons, in-flight downtime, and quiet at-home moments.

   Example. "Triage my work inbox before today's 9am steering check-in."

4. **Theme-implied tool coverage.** Each theme implicitly invokes downstream tools. The {{N}} themes must collectively touch varied tools -- do not converge 15 themes on "search restaurant + book restaurant". Use the capability table below as a sanity-check; do not propose asks no domain supports ("redecorate my living room").

5. **Result-object reachability (commerce / travel / reservation).** A learning session is single-session: no prior turn the agent can rely on to know an existing `order_id` / `trip_id` / `booking_id` / `appointment_id` / `subscription_id`. The current tool set has no `search_orders` / `search_trips` / `search_bookings` / `search_appointments`, so for these three domains the agent cannot fetch an existing result object from a free-text description.

   Counter-example. "Track the bird-seed order I placed for the cardinals." (Commerce; no `search_orders`.)

   Example. "Order another bag of cardinal blend so the feeders do not run out." (Reframed to a NEW `place_order`.)

   This restriction applies ONLY to commerce / travel / reservation. workspace (`list_files`, `search_documents`, `list_trackers`), scheduling (`list_events`, `check_conflicts`), and communication (`search_messages`, `list_labels`) all expose search / list tools, so "update / modify / archive / triage" themes are fine there.

   > Note: this LS restriction does NOT apply to TS -- TS hydrates derived entities from primary ref attributes. LS is first-contact with no hydration.

---

## The 6 available domains

- `commerce` -- shopping / orders / order management / subscriptions / returns
- `reservation` -- restaurant bookings / event tickets / local-service appointments
- `travel` -- flights / hotels / trip planning / ground transport
- `communication` -- email triage / priority / archive & label / drafting replies
- `scheduling` -- calendar events / reminders / meetings / conflicts
- `workspace` -- file organization / documents / archiving / project tracking

## Domain capability table (sanity-check; do NOT name tools in theme text)

{{TOOLS_BRIEF}}

---

## Field spec

- **`domain`** -- pick 1 of 6.
- **`theme_one_line`** -- one English sentence. Carry enough situational color (time, occasion, location hint, mood) for the fill stage to ground the fill realistically. Do not spell out specific param values that should be elicited at runtime (concrete date / time, exact party size, specific cuisine, exact ids) -- those are for the fill stage. Do not name tool functions or domain words verbatim.

  Example. "Book a quiet Saturday dinner spot to celebrate my sister's birthday."

  Counter-example. "Book a Japanese place in the East Village for next Tuesday at 7pm, 4 people." (Leaks cuisine, location, date, party size.)

---

## Input

### Persona structured data

{{STRUCTURED_PERSONA}}

---

## Output

Strict JSON array of {{N}} elements, no code fence and no markdown tags:

```json
[
  {"domain": "<one of 6>", "theme_one_line": "<one English sentence>"},
  ...
]
```

Do NOT output `session_id` / `day_offset` (the caller stamps them) or `situations` / `references_sketch` / `expected_tools` (those moved to the fill stage).

## Self-check (before emitting)

- Exactly {{N}} entries, each with `domain`  in  {commerce, reservation, travel, communication, scheduling, workspace}.
- `theme_one_line` is one English sentence, no tool names, no over-specific param values.
- Sessions interleave across domains and span varied time / mood / location through wording (Principles 1 and 3).
- Themes fit the persona and land within some domain's capability (Principles 2 and 4).
- No commerce / travel / reservation theme starts from an already-existing order / trip / booking / appointment / subscription (Principle 5).
\end{lstlisting}
\end{tcolorbox}

\begin{tcolorbox}[title={Learning-Session Fill (\texttt{learning\_sessions/fill})}]
\begin{lstlisting}
# Learning-Session Fill

The skeleton stage produced a thin skeleton (`domain` + `theme_one_line`). Fill in one complete learning session from that skeleton.

## Core constraints

### Write daily interactions only -- do not reveal preferences

You hold no long-term rule pool. Write only factual scenarios (one specific thing this persona is handing off to the agent right now). Do not hint, in instruction / task_params / references, at "this person generally prefers X" or any cross-task decision direction.

### Lazy-user contract

- **`reason_for_call`** is what the user holds in mind as the overall reason for contacting the agent -- only `user_sim` reads it; the agent never sees it. user_sim uses it to generate the opening message naturally (vague, direction-only).

  `reason_for_call` MUST be vague: convey direction / occasion / mood, but must NOT contain any specific `task_params` value (cuisine, category, location, brand, person name, event title, project name, file / folder name, document type, or any concrete identifier). Generic container words any user might say upfront ("inbox", "calendar", "tracker", "folder", "trip") are fine; specific content words that name *this* user's target are not.

- **`task_params`** is the user's full mental list of specific needs, a flat dict `{field_name: concrete_value}`. Value types: numbers, dates, strings, ids, enum literals, lists -- whatever the field needs. Do NOT add desc / rationale fields; user_sim will paraphrase at runtime.

- Runtime rule for user_sim: when the agent asks about a `task_params` field, paraphrase the value; otherwise reply "I don't know / whatever you think". Repeat the answer if the same field is asked twice.

### Closed-loop traceability (HARD invariants -- every `task_params` key needs a runtime destination)

Every key falls into exactly one of these destinations:

| Destination | Key shape | Value constraint |
|---|---|---|
| A. Direct tool argument | matches a parameter name on a tool the session expects to use | type matches; if enum, value  in  allowed set |
| B. Search filter against refs | matches a `local_env.references[*].attributes` key on the matching ref type | value MUST appear lexically in at least one reference's attribute value |
| C. Reference id selector | ends with `_id` or `_ids` | value(s) MUST exist in `local_env.references[*].id` |
| D. Drafting points | ends with `_points` / `_keywords` / `_bullets` | `list[str]`, each item <= ~80 chars; NOT a pre-authored body |

**Forbidden in `task_params`:**

- **E. Runtime autofill fields** -- `contact_info` / `payment_method` / `shipping_address` / `traveler_info` / `guest_info`. These are auto-injected from PersonaProfile; including them is redundant and will be rejected.
- **F. Server-generated derived IDs** -- `order_id` / `trip_id` / `booking_id` / `appointment_id` / `reservation_id`. LS is first-contact; the user does NOT know server IDs (`ord_xxx`). (TS hydrates derived entities from primary ref attributes; LS does not -- derived IDs in LS refs are also forbidden.) Use a `<noun>_description` field instead (e.g. `order_description: "the recent bird-seed order"`); the agent resolves the actual id via `track_*` / `search_*` at runtime.
- Algorithmic / procedural instructions encoded as values (`selection_rule: "choose the first conflict-free option..."`). Values are *facts*, not *agent algorithms*.
- Output-control booleans (`include_daily_breakdown: true`). Agent output knobs, not user-side facts.

### References pool must support `gold_trajectory`

The trajectory you write must be physically executable against your refs. If the trajectory contains a search step, refs must include >=1 of the corresponding type, or the agent stalls.

### Granularity match

Any string-typed `task_params` value used as a search filter (location / cuisine / category / region) must appear lexically (token-level overlap) in at least one ref attribute. If the user's mental notion is at a different granularity than refs, use the granularity that lexically overlaps refs.

Counter-example. `task_params.location = "Upper Manhattan"` while refs say `location: "Washington Heights"` -- no token overlap, so the env's substring filter returns 0.

Example. `task_params.location = "Washington Heights"` matching `refs[*].attributes.location = "Washington Heights"`.

---

## Inputs

### Skeleton (the skeleton stage has set this; do NOT change domain or invent themes)

```json
{{SKELETON}}
```

### Persona structured data (background, for grounding)

{{STRUCTURED_PERSONA}}

### Domain `{{DOMAIN}}` available tool specs

```json
{{DOMAIN_TOOLS}}
```

### Domain `{{DOMAIN}}` reference object schema

{{REF_SCHEMA}}

---

## Output JSON schema

```json
{
  "reason_for_call": "<<=10 word verb+noun phrase>",
  "task_params": {"<field_1>": <concrete value>, ...},
  "local_env": {
    "references": [
      {"id": "<id with short suffix>", "type": "<valid ref_type>", "attributes": {...}},
      ...
    ]
  },
  "gold_trajectory": [
    {"tool": "<tool name from this domain>", "arguments": {...}},
    ...
  ]
}
```

> `local_env.tools` and `expected_tools`: you do not output these -- the caller fills `tools` with the whole-domain list and derives `expected_tools` from your `gold_trajectory`.

## Field requirements

### `reason_for_call`

Ultra-abstract phrase, <= 10 English words, expressing only "verb + general object". `user_sim` reads it and paraphrases it as the opening message; everything specific stays in `task_params` and is disclosed only when the agent asks.

Strictly forbidden in `reason_for_call`:

- Any `task_params` value, verbatim or as a synonym (cuisine / category / location / brand / event title).
- Any time-of-day / day-of-week / specific occasion / mood adjective ("this weekend", "before today's meeting", "to celebrate").
- Any "head start" that lets the agent skip asking.

Example. "Help me with a dinner reservation."

Counter-example. "Help me book a sushi place in SoHo for two later this week." (Leaks cuisine, location, party size, and date hint.)

Length check: if the draft has more than 10 words, strip everything except verb + general noun phrase.

### `task_params` (3-10 fields, flat key -> value)

- Include ALL the information the agent needs to complete the task.
- Values are concrete (numbers, dates, place names, headcounts, ids, enum values) -- no placeholders.
- Every field must be answerable by user_sim when asked. Since the instruction reveals nothing, every field is obtained by the agent asking or searching.
- Every (key, value) satisfies one of destinations A--D in the traceability table above.
- Do not stuff agent output text into a value: no full body for `reply_body` / `document_content` -- use a `*_points` drafting list.
- **Executability**: think clearly about the last action tool the agent calls (`book_` / `place_` / `send_` / `create_` / `cancel_` / ...). Its required parameters must come from a value in `task_params`, an id from `local_env.references`, or a runtime auto-injected identity field (`contact_info` / `traveler_info` / `guest_info` / `shipping_address` / `payment_method` -- these MUST NOT appear in task_params).

### `gold_trajectory` (oracle solution path -- 2-5 steps)

The sequence of tool calls an agent would make if it knew all `task_params` upfront (no asking). Each step is `{"tool": <name>, "arguments": <dict>}`. The caller derives unique tool names from this trajectory.

Hard requirements:

- Each step's `tool` is in this domain's tool list (`{{DOMAIN_TOOLS}}`).
- **Every concrete argument value** (required and optional, search filters and write payloads) must come from one of:
  1. A value in `task_params` (verbatim).
  2. An id from `local_env.references[*].id`.
  3. A placeholder `<authored from <field>[, <field>...]>` for long-text drafting args (every referenced `<field>` must exist in `task_params`).
  4. A runtime auto-injected identity field, OMITTED from arguments (env auto-fills).
  5. A recoverable broad-list default for narrowing filters the agent could safely omit: `search_messages.folder  in  {inbox, sent, drafts}` (omitting returns all non-archived messages); `list_events.calendar` (omitting returns all calendars). NOT recoverable: `search_messages.folder = "archive"` (archived messages are hidden by default -- this must come from task_params).
  6. An agent-side control flag (`sort_by`, `limit`, `language` defaults, output-shaping booleans, `field` / `update_action` selectors). These are how the agent invokes the tool, not user-side facts, and need not appear in task_params.
- Anything else (a `doc_type`, `cuisine`, `location`, `sender`, `priority`, `title`, `start_time`, `participants` invented only in `gold_trajectory`) is oracle-only knowledge user_sim cannot disclose -- fail.
- Every `search_*` / `list_*` step must be answerable by your refs (>=1 ref of the matching type satisfying the step's filters).
- Every id used as an argument must exist in `local_env.references` with the right type.
- The last step is typically the action tool. Trajectory length 2-5 (search -> optional compare -> action).

For long-text drafting args (e.g. `body` on `draft_message`, `content` on `create_document`), use a placeholder like `"<authored from reply_points>"` -- do not compose the full text.

**Examples** (replace `<task_params.X>` with the literal values you wrote):

```json
// reservation
[
  {"tool": "search_restaurants", "arguments": {"cuisine": "<task_params.cuisine>", "location": "<task_params.location>", "sort_by": "rating"}},
  {"tool": "book_restaurant",   "arguments": {"restaurant_id": "<gold ref id>", "date_time": "<task_params.date_time>", "party_size": 2}}
]

// communication
[
  {"tool": "search_messages",      "arguments": {"folder": "inbox", "sender": "<task_params.target_sender>"}},
  {"tool": "set_message_priority", "arguments": {"message_ids": ["<msg id>"], "priority": "high"}},
  {"tool": "draft_message",        "arguments": {"thread_id": "<thread id>", "language": "en", "body": "<authored from reply_points>"}}
]
```

### `local_env.references` (3-6 entries)

- Each ref: `{id, type, attributes}`. `type` must be valid in `{{REF_SCHEMA}}` -- do not invent. `attributes` follow the schema contract. `id` follows the schema's recommended prefix (`rst_`, `prod_`, `msg_`) with a numeric suffix (e.g. `rst_momoya_01`).
- Cover every search-style tool in `gold_trajectory`: if a step is `search_restaurants`, refs must include >=1 `restaurant`; same for `search_destinations` -> `destination`, `list_events` -> `calendar_event`.
- Diverse and natural distribution -- avoid 4 near-identical items.
- References hold only pre-existing candidate objects available before the agent starts. Do not put result objects (orders already placed, bookings already made) in references; use a descriptive field like `order_description: "the recent art-supply order"` and let the agent fetch via `track_*` / `search_*`.
- Derived IDs in ref attributes are LS-forbidden. REF_SCHEMA's `derived:` sections describe TS-only env hydration; LS is interactive first-contact with no hydration, so do NOT put `order_id` / `trip_id` / `booking_id` / `reservation_id` / `appointment_id` in any ref's attributes.

---

## Domain-specific atomic constraints

### scheduling -- `title` + `start_time` + `end_time` must come from the SAME calendar_event ref

If `task_params` includes any of `title` / `start_time` / `end_time` for an existing event, they must jointly match a single `calendar_event` ref's attributes. Treat the triplet as one atomic write -- change one, sync the other two. Picking the title from one ref and the time from another (or made-up time) is the most common scheduling-domain failure.

### workspace -- reference doc_type uses a dedicated `task_params` field

When `gold_trajectory.search_documents` uses a `doc_type` different from `task_params.doc_type` (the agent looks at an existing template / reference before creating the user's actual document), declare the search type under one of: `source_doc_type` / `template_doc_type` / `reference_doc_type`. If multiple reference types are consulted, use a distinct field for each.

Example. `task_params: {doc_type: "note", source_doc_type: "template", ...}` -> gold: `search_documents(doc_type="template")` then `create_document(doc_type="note")`.

Counter-example. `task_params: {doc_type: "note", ...}` -> gold: `search_documents(doc_type="template")` -- no `task_params` field carries the reference type, so user_sim cannot answer if asked.

---

JSON only -- no markdown code fence, no other text.
\end{lstlisting}
\end{tcolorbox}

\subsection{Agent Runtime Prompts}
\label{app:agent_runtime_prompts}

We show runtime prompt templates with placeholders for dynamic context. The domain-specific policy is listed separately in Appendix~\ref{app:domain_policy_prompts}; long \texttt{<context>} contents are elided.

\begin{tcolorbox}[title={Agent Learning-Session Prompt Template}]
\begin{lstlisting}
<instructions>
You are the user's long-term personal assistant. You serve this same user
across many tasks over time.

## Execution Conventions
- Any information must be obtained through search / list / track tools first
  -- never fabricate IDs.
- ID parameters must come from tool results, not from names or keywords. See
  `<policy>` for the ID discovery path in each domain.
- Identity fields are auto-filled from the user's account defaults -- omit
  them even when the tool schema marks them optional, and never block on
  them: `shipping_address`, `payment_method`, `contact_info`,
  `traveler_info`, `guest_info`.
</instructions>

{DOMAIN_POLICY}

<context>
{omitted: cross-session context block, if non-empty}
</context>

<session_mode>
The user is online and will respond to your messages.

## Protocol
- Complete the user's request and tell the user when done.
- When you lack key inputs to complete the current task, ask the user rather
  than calling tools with guessed values.
- If you can't make progress -- empty search/list results after one or two
  reasonable attempts, or any other blocker -- ask the user rather than
  looping on tools.
</session_mode>

<send_protocol>
The user only hears you through the `send_to_user` tool. Plain assistant text
outside this tool is internal reasoning only -- the user will not see it.
</send_protocol>

{VARIANT_ADDITION}
\end{lstlisting}
\end{tcolorbox}

\begin{tcolorbox}[title={Agent Test-Session Prompt Template}]
\begin{lstlisting}
<instructions>
You are the user's long-term personal assistant. You serve this same user
across many tasks over time.

## Execution Conventions
- Any information must be obtained through search / list / track tools first
  -- never fabricate IDs.
- ID parameters must come from tool results, not from names or keywords. See
  `<policy>` for the ID discovery path in each domain.
- Identity fields are auto-filled from the user's account defaults -- omit
  them even when the tool schema marks them optional, and never block on
  them: `shipping_address`, `payment_method`, `contact_info`,
  `traveler_info`, `guest_info`.
</instructions>

{DOMAIN_POLICY}

<context>
{omitted: raw context block or oracle target-rule statement, if non-empty}
</context>

<session_mode>
The user is offline.

Complete the user's request independently.

## Termination
- When the task is complete, or when no path forward remains: must send a
  single `finish_session` tool call with no other content.
- On tool errors, try different parameters or a different tool before giving
  up.
</session_mode>
\end{lstlisting}
\end{tcolorbox}

\begin{tcolorbox}[title={Variant-Specific Learning Additions}]
\begin{lstlisting}
--- default ---
{empty}

--- atr ---
<standing_rule>
A standing rule:
- is a long-term user preference;
- pins how you should act on a recurring decision;
- holds across the user's future tasks;
- is not a specific instance from the task.
</standing_rule>

<standing_rule_acquisition>
You may ask about a standing rule of theirs that might prove useful in
serving them later.

Asking costs the user's patience (cost); a question that matches a real
rule of theirs reduces future errors (benefit).
</standing_rule_acquisition>

--- always_ask ---
<standing_rule>
A standing rule:
- is a long-term user preference;
- pins how you should act on a recurring decision;
- holds across the user's future tasks;
- is not a specific instance from the task.
</standing_rule>

<standing_rule_acquisition_required>
After you have largely finished the user's current task, ask exactly one
question about a standing rule of theirs that may prove useful in serving
them later. Do not ask more than one such question in this conversation.
</standing_rule_acquisition_required>
\end{lstlisting}
\end{tcolorbox}

\begin{tcolorbox}[title={Oracle Context Injection}]
\begin{lstlisting}
The oracle variant skips learning sessions. In each test session, the
same test-session prompt template is used, with the context block containing
only the canonical answer for the target rule:

<context>
[Prior statement 1]
[user]: {canonical answer for the target rule}
</context>
\end{lstlisting}
\end{tcolorbox}

\subsection{Agent Domain Policy Prompts}
\label{app:domain_policy_prompts}

\begin{tcolorbox}[title={Agent Domain Policy Prompts}]
\begin{lstlisting}
--- domain_commerce ---
<policy domain="commerce">
## ID discovery paths
- `subscription_id` <- `review_recurring_charges` (no search_subscriptions tool)
- `order_id` <- `search_products` result attributes, previous `place_order`
  return, or instruction (no search_orders tool)

</policy>

--- domain_reservation ---
<policy domain="reservation">
## ID discovery paths
- `reservation_id` / `booking_id` / `appointment_id` <- previous `book_*`
  return or instruction (no search_reservations / search_bookings /
  search_appointments tool)

## Shape constraints
- `track_reservation_updates`: exactly one of `reservation_id` /
  `booking_id` / `appointment_id`

</policy>

--- domain_travel ---
<policy domain="travel">
## ID discovery paths
- `plan_trip.selected_stop_ids` /
  `replan_trip.selected_alternative_stop_ids` <- `search_trip_stops`
- `booking_id` <- previous `book_*` return
- `trip_id` <- previous `plan_trip` return

## Shape constraints
- `track_trip_updates`: exactly one of `trip_id` and `booking_id`

</policy>

--- domain_communication ---
<policy domain="communication">
## ID discovery paths
- `label_ids` <- `list_labels`
- `message_ids` <- `search_messages`
- `thread_id` (for replying) <- `search_messages`
- `folder_id` (optional) <- `list_message_folders`
- `draft_id` <- previous `draft_message` return

## Shape constraints
- `draft_message` / `send_message`: at least one of `recipient` and `thread_id`

</policy>

--- domain_scheduling ---
<policy domain="scheduling">
## ID discovery paths
- `event_id` <- `list_events`, `check_conflicts`, or previous `create_event`
  return
- `target_id` (for `set_reminder` / `track_event_updates` /
  `find_alternative_slots`) <- same sources as `event_id`

</policy>

--- domain_workspace ---
<policy domain="workspace">
## ID discovery paths
- `file_ids` <- `list_files`
- `folder_id` <- `list_file_folders`
- `tracker_id` <- `list_trackers`
- `document_id` <- `search_documents`
- `create_document`: no ID input needed

</policy>
\end{lstlisting}
\end{tcolorbox}

\subsection{User Simulator Prompts}
\label{app:user_sim_prompts}

\begin{tcolorbox}[title={User Simulator Opening Prompt}]
\begin{lstlisting}
--- system ---
You play the user, having just contacted the task assistant. Send the first message to open the conversation.

- Write 1-2 conversational sentences.
- State just enough immediate task context for the assistant to begin
  (for example, "haircut in Decatur" or "flight from Atlanta to Las
  Vegas").
- Leave follow-up details, choices, and execution specifics for the
  assistant to ask about or look up.
- Do not reveal hidden/reference ids, exact selected targets, exact
  titles/names, destination folder/list/project names, drafted content
  points, or long-term preferences.

Output only the opening line -- no explanation, quotes, or prefix.

## Intent
{learning_session.reason_for_call}

## Opening hint        (optional; sanitized from the first gold step)
- `{arg_key}`: {arg_value}
...

--- user ---
Start the conversation.
\end{lstlisting}
\end{tcolorbox}

\begin{tcolorbox}[title={User Simulator Reply Prompt}]
\begin{lstlisting}
--- system ---
You play the user, talking with the task assistant.

You have one tool: `mark_task_complete` -- call it to signal the session is done.

Your job is to answer task questions from your `## Requirements` and help the assistant complete the current task.

## How to answer
- If the assistant asks about something covered in your Requirements, answer from there -- match the way it asked (one detail if it asked for one; several together if it asked for several).
- If the assistant asks about anything not in your Requirements, you don't have an answer to give. Say something like "no strong preference there -- your call" / "whatever works" / "up to you". Even when framed as a binary choice ("A or B?"), do not pick A or B; picking either side counts as making up an answer you don't have.
- If the assistant asks for help or seems stuck, use your `## Guided path` as a reference and tell it what to do next in everyday language.

## Style
- Speak only when prompted; don't volunteer information beyond what's directly relevant.
- Use everyday language; describe content, not tool names, field names, or parameter names.

## Ending the session
Stay within your stated intent. The conversation shows `[invoked tool_name(...) -> ok]` markers for actions the assistant has completed; the final action in your `## Guided path` showing such a marker is your signal that the core task is done.

Keep replying normally while the assistant is collecting details, asking questions, performing the task, reporting progress, or waiting for any user-visible answer from you.

After the task is complete, close the conversation naturally and gradually move it toward a normal ending. Keep replying while the assistant is still responding to that closing exchange.

Call `mark_task_complete` (in its own turn, with no text) only after the assistant has acknowledged the closing, said goodbye, or made only a generic final offer of further help.

If the task cannot be completed, terminate the same way via `mark_task_complete`.

## User background
{persona runtime narrative}

## Intent
{learning_session.reason_for_call}

## Requirements
- `{task_param_key}`: {task_param_value}
  {optional referenced-object attributes}

## Guided path
1. {tool_name}({arguments})
...

--- messages ---
{prior conversation, with agent messages as user-role messages and user-simulator replies as assistant-role messages}

--- user ---
{agent_text}

--- optional user message on Router-positive turns ---
{rule_answer_hook}
\end{lstlisting}
\end{tcolorbox}

\subsection{Router and Classifier Prompts}
\label{app:router_classifier_prompts}

The placeholder \verb|{{FEW_SHOT_EXAMPLES}}| is replaced by the active shot bank before the scaffold LLM call.

\begin{tcolorbox}[title=Router Prompt]
\begin{lstlisting}
You decide whether one agent message asks the user about a standing rule.

A standing rule is a user preference, habit, or default that would shape
the agent's behavior on a different future task.

## Input

- `<reason>`: the agent's free-form intent for this turn. Treat it as
  the agent's own statement of why it is speaking.
- `<output>`: what the agent actually said to the user.

## Decision

Return `is_strict_rule_question = true` when `<output>` contains a question
the agent itself asks about a standing rule. Use `<reason>` as supporting
evidence, but do not require it to explicitly say "standing rule" if the
user-visible output clearly asks for a future/default behavior.

Return `false` when `<reason>` or `<output>` only applies, restates, or
confirms a rule the user has already stated.

Recall of past rules, status updates, and quoted draft content are
context, not asks.

Return `false` in any other case.

## rule_question_span

When `true`, copy one complete verbatim standing-rule ask sentence from
`<output>`. Direct questions such as `would you ...?` and soft/indirect asks
such as `let me know if you'd like ...` both count when they ask about a
future/default behavior. Include sentence-level lead-ins or framing that belong
to that ask, such as `for future ...`, `For recurring ...`, `Quick question for
the future:`, or `Since ..., would you ...`. Do not paraphrase or include
previous independent status, recall, draft, or current-task sentences. Preserve
the original capitalization, quotes, punctuation, and line breaks exactly.

When `false`, set to `null`.

{{FEW_SHOT_EXAMPLES}}

## Output format

Return exactly one JSON object, nothing else:

```json
{
  "is_strict_rule_question": false,
  "rule_question_span": null
}
```
\end{lstlisting}
\end{tcolorbox}

\begin{tcolorbox}[title=Classifier Prompt]
\begin{lstlisting}
You are a rule selector.

The assistant just asked the user a question to learn one of the user's
standing rules. From a pool of candidate rules, pick the one the user
would cite when answering the question -- or return `null` if no rule
in the pool fits.

## Inputs

- `question`: what the assistant asked, posing a recurring preference
  or default.
- `rule_pool`: a list of `{rule_id, statement}` pairs. Each `statement`
  is a first-person rule in one of the forms "I always X for Y" /
  "When Y, I do Z" / "For X, I prefer Y".

## What counts as a fit

A rule fits the question only when BOTH conditions hold:

1. **Trigger covers**: the rule's trigger context covers the situation
   the question describes.
2. **Behavior answers**: the rule's preferred behavior directly answers
   the action, value, or choice the question proposes.

If either condition is weak or unclear, return `null`.

The question's wording does not need to match the rule's word-for-word.
But the rule's content, applied as the user's answer, must resolve what
the question is asking -- not merely live in the same neighborhood.

Two recurring patterns to recognize:

- **Specific instance of a broader rule**: the question may name a
  specific entity, keyword, or narrower scenario that falls within the
  rule's broader trigger category. This is still a fit only when the
  rule's behavior directly answers the question. If the question asks
  about a different dimension than what the rule pins, return `null` --
  even when trigger keywords overlap.

- **Choice question**: the question may list options. If one option is
  the rule's preferred behavior, this is a fit.

## When to return `null`

Default to `null` unless a rule clearly satisfies both conditions above.
Specifically return `null` when:

- No rule's trigger context covers the situation in the question.
- A rule's trigger overlaps in domain or keyword, but the rule's
  behavior is on a different dimension or axis than the question's
  proposal.
- The question is about a one-off action in this task, not a recurring
  preference.
- The question asks for an implementation detail of a rule the user
  has already stated earlier.
- Several rules are broadly related, but none of them, on its own,
  would resolve the question.

## Tie-break

If two or more rules look like fits, pick the one whose trigger and
behavior align most directly with the question. Prefer a same-level
match over a "specific instance" match.

{{FEW_SHOT_EXAMPLES}}

## Output

Strict JSON, no extra text:

{
  "rule_id": "<rule_id from rule_pool>" | null,
  "reasoning": "<one sentence: why this rule fits, or why no rule fits>"
}
\end{lstlisting}
\end{tcolorbox}

\begin{tcolorbox}[title={Router User Message}]
\begin{lstlisting}
<reason>{send_to_user.reason}</reason>
<output>{send_to_user.output}</output>
\end{lstlisting}
\end{tcolorbox}

\begin{tcolorbox}[title={Classifier User Message}]
\begin{lstlisting}
## Rule pool
- rule_id={rule_id}
  user statement: {rule.canonical_answer}
...

## Rule question span
{router.rule_question_span}
\end{lstlisting}
\end{tcolorbox}

\begin{tcolorbox}[title={Transient Scaffold Hooks}]
\begin{lstlisting}
--- send_to_user repair hook ---
<scaffolding_note>
Your last assistant turn was not delivered to the user because it didn't go
through the send_to_user tool. Please re-emit the user-facing content via
send_to_user(output, reason).
</scaffolding_note>

--- rule-answer hook ---
<rule_answer>
Even though this question is not covered by your Requirements and would
normally get a no-preference deflect, for the question
"{rule_question_span}" in the assistant's last message your answer is
<RULE_ANSWER>. Use this exact token once in your reply where the answer
would naturally fit; don't write a substantive answer yourself. Reply to
anything else in the assistant's message as you normally would.
</rule_answer>
\end{lstlisting}
\end{tcolorbox}
}